\documentclass[10pt,twocolumn,letterpaper]{article}

\usepackage[pagenumbers]{iccv} 

\usepackage{multirow}
\usepackage{makecell}
\usepackage{amssymb}

\usepackage{tabularray}

\definecolor{iccvblue}{rgb}{0.21,0.49,0.74}
\usepackage[pagebackref,breaklinks,colorlinks,allcolors=iccvblue]{hyperref}

\def\paperID{***} 
\def\confName{ICCV}
\def\confYear{2025}

\title{ELIP: Enhanced Visual-Language Foundation Models for Image Retrieval}

\author{%
  Guanqi Zhan$^{1*}$, Yuanpei Liu$^{2*}$,
  Kai Han$^2$, Weidi Xie$^{1,3}$, Andrew Zisserman$^1$\\
    $^1$VGG, University of Oxford\quad\quad
    $^2$The University of Hong Kong \quad\quad
    $^3$Shanghai Jiao Tong University\\
  \texttt{\{guanqi,weidi,az\}@robots.ox.ac.uk} \\
  \texttt{ypliu0@connect.hku.hk}~~~~
  \texttt{kaihanx@hku.hk} \\
}

\begin{document}

\twocolumn[{
    \vspace{-30pt}
    \renewcommand\twocolumn[1][]{#1}
    \maketitle
    \centering
    \vspace{-10pt}
 \includegraphics[height=0.45\linewidth]{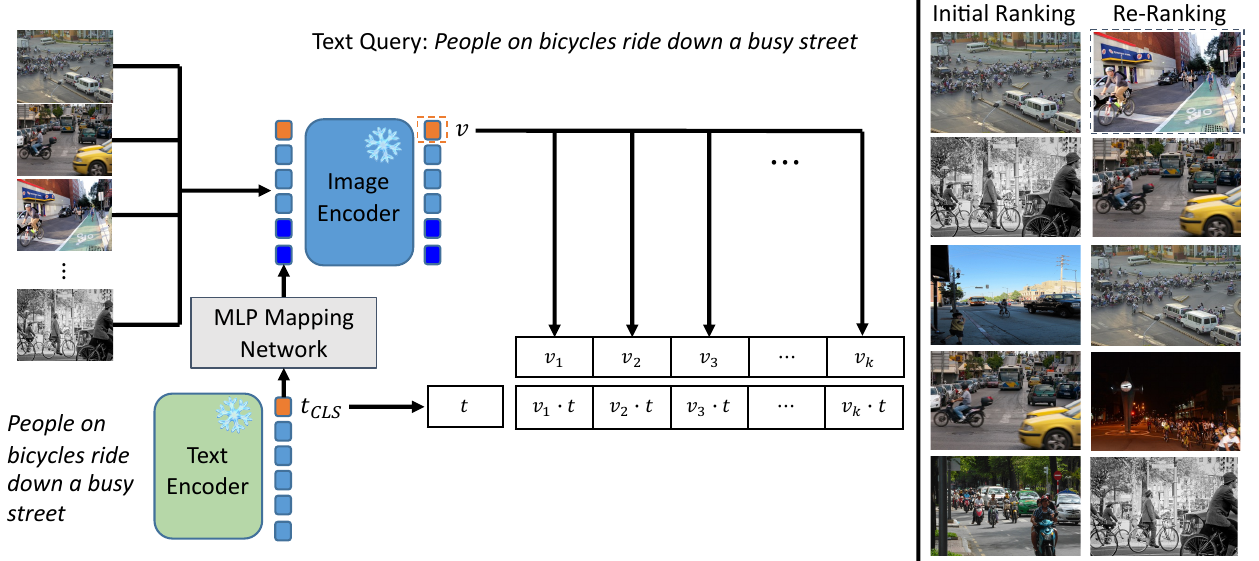}   
 \vspace{-8mm}
   \captionof{figure}{
\textbf{The ELIP architecture.}
{\em Left}: We propose a novel architecture that can be applied to pre-trained and frozen vision-language foundation models, such as CLIP, SigLIP, SigLIP-2 and BLIP-2, to enhance their text-to-image retrieval performance. 
The \emph{key idea} is to use the text query to define a set of visual prompt vectors that are incorporated into the image encoder to make it aware
of the query when generating the embedding. An MLP maps from the text space to the visual space of the input to the ViT encoder. The architecture is lightweight, and our data curation strategies enable efficient and effective training with limited resources.
{\em Right}: In this retrieval example from the COCO benchmark, the top-$k$ ($k$=100) images are re-ranked by our ELIP model for the text query: `People on bicycles ride down a busy street'. The ground truth image matching the query is not in the top-5 ranked images in the initial CLIP ranking, but is ranked top-1 (highlighted in the dashed box) by the re-ranking. 
   }
    \label{fig:teaser}
    \vspace{20pt}
    }
    ]

\def\thefootnote{*}\footnotetext{Equal contribution.}\def\thefootnote{\arabic{footnote}}

\begin{abstract}

The objective in this paper is to improve the performance of text-to-image retrieval. To this end, we introduce a new framework that can boost the performance of large-scale pre-trained vision-language models, so that they can be used for text-to-image re-ranking. The approach, \textbf{Enhanced Language-Image Pre-training} (\textbf{ELIP}), uses the text query, via a simple MLP mapping network, to predict a set of visual prompts to condition the ViT image encoding. ELIP can easily be applied to the commonly used CLIP, SigLIP and BLIP-2 networks. 
To train the architecture with limited computing resources, 
we develop a `student friendly' \emph{best practice}, involving global hard sample mining, and curation of a large-scale dataset. On the evaluation side, we set up two new out-of-distribution (OOD) benchmarks, \emph{Occluded COCO} and \emph{ImageNet-R}, to assess the zero-shot generalisation of the models to different domains. 
The results demonstrate that ELIP {\em significantly} boosts CLIP/SigLIP/SigLIP-2 text-to-image retrieval performance and outperforms  BLIP-2  on several benchmarks, as well as providing an easy means to adapt to OOD datasets.

\end{abstract}    
\section{Introduction}
\label{sec:intro}
This paper considers the problem of text-to-image retrieval, 
that aims to rank image instances based on their relevance to a text query. 
Effective retrieval generally includes two stages: the first stage provides an initial ranking in a fast and efficient manner, while the second stage—referred to as re-ranking--refines this ranking by re-computing the relevance scores between the text query and each of the top-ranked candidates with a more expensive model.

Recent advances in text-to-image retrieval have primarily focused on the first stage. Notable models, such as CLIP~\cite{radford2021learning} and ALIGN~\cite{jia2021scaling}, leverage contrastive learning~\cite{oord2018representation} on large-scale image-text pairs to learn joint representations, demonstrating impressive generalization capabilities for cross-modal retrieval tasks.

Our primary contribution here focuses on the second stage of the retrieval pipeline, namely, the re-ranking. Specifically, our goal is to enhance the performance of off-the-shelf vision-language foundation models, so that they can be re-purposed for re-ranking the top-$k$ candidates from the fast retrieval process. 
The approach we develop, termed as {\em Enhanced Language-Image Pre-training} ({\bf ELIP}), requires only a few trainable parameters, and the training can be conducted efficiently with `student friendly' resources and data. 
We demonstrate that ELIP can boost the performance of the pre-trained CLIP~\cite{radford2021learning}, SigLIP~\cite{zhai2023siglip}, SigLIP-2~\cite{tschannen2025siglip2} and BLIP-2~\cite{li2023blip} for cross-modal retrieval.

To achieve this goal, we first introduce a lightweight, text-guided visual prompting module. As illustrated in Figure~\ref{fig:teaser}, a query text is mapped to a set of visual prompt vectors~\cite{jia2022visual}, that are then concatenated with the \texttt{[CLS]} and patch embeddings of the image encoder. These augmented embeddings are then passed into the frozen vision encoder to re-compute the image representation. The resulting image embedding is aware of the text conditioning and this enhances its performance in re-ranking.

As a second contribution, we address the two major challenges in training large vision-language models: first, data size -- to enable a strong generalisation capability it is necessary to train on millions or billions of images, but this is expensive; second, the batch size -- to enhance the model's discriminative capability it is important to train at a large batch size, but that requires a large number of GPUs. 
We develop here a best practice by introducing strategies to select and curate a training dataset with maximum information, and group hard samples together in a batch to make training with small batch size effective.

To assess the re-ranking performance of our proposed ELIP models, 
we experiment on the standard COCO~\cite{lin2014microsoft} and Flickr30k~\cite{plummer2015flickr30k} text-to-image retrieval benchmarks. 
As a further challenge, we also evaluate the generalisation of the ELIP-boosted models on out-of-distribution domains. To do so, we repurpose the Occluded COCO~\cite{lee2022instance} and ImageNet-R~\cite{hendrycks2021many} datasets to be used for text-to-image retrieval benchmarks.

In summary, we have made the following four contributions:
\emph{First}, we propose a novel architecture to improve text-based image retrieval on large pre-trained vision-language models, including the most popular CLIP/SigLIP architectures 
and the state-of-the-art BLIP-2 architecture.
\emph{Second}, we propose a best practice for training our architecture efficiently with limited resources.
\emph{Third}, to evaluate the generalisation capability of text-to-image retrieval models to different out-of-distribution domains, we set up two new benchmarks of text-to-image retrieval, \emph{Occluded COCO} and \emph{ImageNet-R}. 
\emph{Fourth}, and most significantly, we demonstrate that ELIP  {\em substantially} improves the image retrieval performance of CLIP and SigLIP architectures, and outperforms the state-of-the-art BLIP-2 architecture. Furthermore, it provides an efficient method
to adapt these architectures to OOD datasets, again giving a tremendous boost with CLIP, SigLIP, SigLIP-2, and BLIP-2.  

\section{Related Work}
\label{sec:related_work}

\noindent \textbf{Text-to-Image Retrieval}  
is a fundamental and much researched task in cross-modal learning~\cite{lee2018stacked,chen2020imram,zhang2022negative,radford2021learning,li2021align,chen2020uniter,wang2023image,yu2022coca,chen2024internvl,chen2021learning,chen2023vilem,chun2021probabilistic,diao2021similarity,engilberge2018deep,gu2018look,huang2017instance,ji2019saliency,karpathy2014deep,kim2023improving,li2019visual,liu2019focus,liu2020graph,song2019polysemous,thomas2020preserving,wang2020consensus,wang2018learning,wang2019camp,wang2023multilateral,wei2020universal,wei2020multi,yan2021discrete,zeng2022learning,zhang2022show,zhang2020context,zhang2018deep,zheng2019towards,vendrow2024inquire,kordopatis2025ilias}.
Large vision language models such as CLIP~\cite{radford2021learning,openclip}, ALIGN~\cite{jia2021scaling}, BLIP-2~\cite{li2023blip}, SigLIP~\cite{zhai2023siglip} and SigLIP-2~\cite{tschannen2025siglip2} that have powerful zero-shot capabilities have now become the de facto method for open-set text-based image retrieval.
The most recent work~\cite{sogi2024object} gives a slight improvement over BLIP-2 by incorporating the output of an object detector or annotations of detection bounding boxes. This succeeds in overcoming the failure cases where small but semantically important objects in an image are not properly understood by the model. We compare to this model and show superior performance.

\noindent \textbf{CIR and Universal Retrieval.}  
In composed image
retrieval~(CIR)~\cite{liu2023CIR,gu2023compodiff,baldrati2023zero,ventura2024covr}, 
the query is specified by a composition of an image and text, with the
text specifying how the image should be changed. For example, the query image may be of a dog lying down, and the query text may be `playing with a ball'. This composed query defines
the target image to be retrieved from the gallery. This differs from our task, where the query is specified
only by text, and the text alone defines the target image to be retrieved from the gallery.  A more general setting is `universal retrieval'~\cite{wei2023uniir,liu2025LamRA} 
where the query can be a combination of image,
text, and instruction;  and the target can be image alone, text alone, or image and text.

\vspace{2pt}
\noindent \textbf{Post-Retrieval Re-ranking.}
For single modality image retrieval, where the query is an image, there has been a series of works that have re-ranked the top-$k$ images from an initial ranking via classical computer vision algorithms, such as `query expansion', `geometric verification', or a combination of the two~\cite{jegou2008hamming,philbin2007object,chum2011total,chum2007total,tolias2014visual,Arandjelovic12}, as well as via learning-based algorithms~\cite{cao2020unifying,hausler2021patch,el2021training,tan2021instance,Bhalgat23}. 
Re-ranking algorithms have been relatively less explored in text-to-image retrieval~\cite{yanagi2019text,qu2023learnable,long2024cfir}. \cite{miech2021thinking} introduced a method for computing the similarity score between an image and a text query by estimating the log-likelihood of the text conditioned on the image. While this approach has demonstrated strong performance, it remains computationally expensive both during training and inference, making it a {\em slow} process.
Our paper also focuses on the re-ranking stage -- developing a more powerful version of visual-language foundation models to give a better ranking of images that are  hard to distinguish by the original retrieval model.

\vspace{2pt}
\noindent \textbf{Multi-Modal Datasets.} 
To obtain multi-modal foundation models with a strong capability of generalisation, it is important to train them on large-scale multi-modal datasets.
Therefore, in recent years, there has been a significant increase in the number and scale of multimodal vision-language datasets that provide image-text pairs, such as
COCO~\cite{lin2014microsoft}, SBU~\cite{ordonez2011im2text}, Conceptual Captions~\cite{sharma2018conceptual}, LAION~\cite{schuhmann2022laion}, DataComp~\cite{gadre2024datacomp}. The increase in the size of multi-modal datasets enables the training of more powerful visual-language foundation models. More recently, DataCompDR~\cite{vasu2024mobileclip} utilises prior knowledge from large-scale pre-trained image captioning models to generate synthetic captions for DataComp images which results in less noisy captions than the datasets collected from the web, such as the original DataComp dataset. In our paper, we have experimented with training our model using Conceptual Captions~\cite{sharma2018conceptual} and DataCompDR~\cite{vasu2024mobileclip}.

\vspace{2pt}
\noindent \textbf{Multi-Modal Data Curation.} 
It is essential to conduct data curation for multi-modal datasets as it enables more efficient and effective training, especially in situations where the resources are limited.
There have been continuous efforts in data curation, such as offline example-level data pruning~\cite{gadre2024datacomp,kakaobrain2022coyo-700m,changpinyo2021conceptual,jia2021scaling,fang2023data,xu2023demystifying,hessel2021clipscore,mahmoud2024sieve}, offline cluster-level data pruning~\cite{abbas2023semdedup,abbas2024effective,sorscher2022beyond,campbell2018bayesian,har2004coresets}, and online data curation with model-based scoring~\cite{evans2023bad,lin2024rho,loshchilov2015online,mindermann2022prioritized}. 
The most recent work, JEST~\cite{evans2024data}, utilises a pair of learner model and reference model to select batches of data that are learnable by the model but have not yet been learned. This inspired us to select the most efficient batches to train our architecture on BLIP-2. Another series of works related to us is hard negative mining, which has been both explored in classical metric learning~\cite{bucher2016hard,harwood2017smart,mishchuk2017working,simo2015discriminative,wu2017sampling,xuan2020hard} and modern contrastive learning~\cite{robinson2020contrastive,tian2021divide}.

\section{Preliminaries}
\label{sec:preliminary}

\noindent \textbf{Re-Ranking in Image Retrieval.} 
Given an input query, the goal of a retrieval system is to rank all instances in a dataset $\Omega = \{I_1, \dots, I_n \}$, based on their relevance to the query. 
In the case of text-to-image retrieval, the query is specified by text~($T$), and the ideal outcome is a set~($\hat{\Omega}$), with the relevant images being ranked higher than those that are not. In general, an effective retrieval system proceeds in two stages: the first stage provides an initial ranking in a fast and efficient manner, while the second stage—referred to as re-ranking--refines this ranking by recomputing the relevance scores between the text query and each of the top-$k$ ranked candidates with a more powerful (and usually more expensive) ranking model. 
The $k$ is selected such that in general there is a high recall for all the relevant images. In this paper, our novelty lies on the second stage, 
that aims to re-rank the top-$k$ candidates from the first stage results. 

\vspace{3pt}
\noindent \textbf{Visual Prompt Tuning~(VPT)}~\cite{jia2022visual} is a method of enhancing the ViT image encoder by inserting additional learnable prompts into the transformer layers. It enables efficient adaptation of ViT, requiring  only the few parameters of the learnable prompts to be trained. VPT has two different variants -- \emph{VPT-Shallow} and \emph{VPT-Deep}. \emph{VPT-Shallow} only inserts the additional visual prompts into the first Transformer layer, whereas for \emph{VPT-Deep}, prompts are introduced at every transformer layer's input space. We insert our generated set of visual prompt vector into the first transformer layer of ViT, which is similar to \emph{VPT-Shallow}.

\section{The ELIP Architecture}
\label{sec:mapping_network}

In this section, we describe the ELIP text-to-visual prompt mapping network, 
that can be efficiently applied to adapt the commonly used CLIP/SigLIP architectures as well as the more sophisticated BLIP-2 architectures for re-ranking. We first introduce the architecture of the network in Section~\ref{sec:network}, and the training/inference strategy in Sections~\ref{sec:arch-clip} and~\ref{sec:arch-blip} respectively. 
We refer to the network applied to CLIP as \emph{ELIP-C}, applied to SigLIP/SigLIP-2 as \emph{ELIP-S/ELIP-S-2}, and applied to BLIP-2 as \emph{ELIP-B}.

\begin{figure}[t]
	\centering
\includegraphics[height=0.45\linewidth]{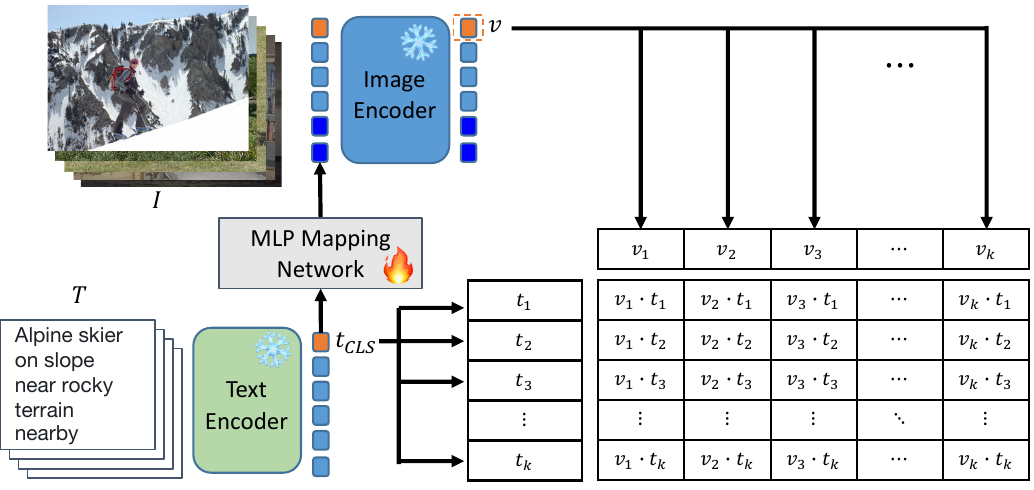}
\vspace{-2mm}
	\caption{
\textbf{Architecture of ELIP-C / ELIP-S.}  
At training time, a batch of text-image pairs is fed into the architecture.
The text feature is mapped to the visual embedding space as a set of prompt vectors via the MLP mapping network and then guides the encoding of the image feature. 
We use color coding for the {\color{orange} \texttt{[CLS]} token}, 
{\color{cyan} patch tokens}, and {\color{blue} generated visual tokens} from text. The architecture is trained with InfoNCE loss (for ELIP-C) and Sigmoid loss (for ELIP-S/ELIP-S-2), to align the text feature with the corresponding re-computed image feature.} 
\vspace{-3mm}
\label{fig:arch-1}
\end{figure}

\subsection{Text-Guided MLP Mapping Network}
\label{sec:network}

\begin{figure}[t]
\centering
\includegraphics[height=0.45\linewidth]{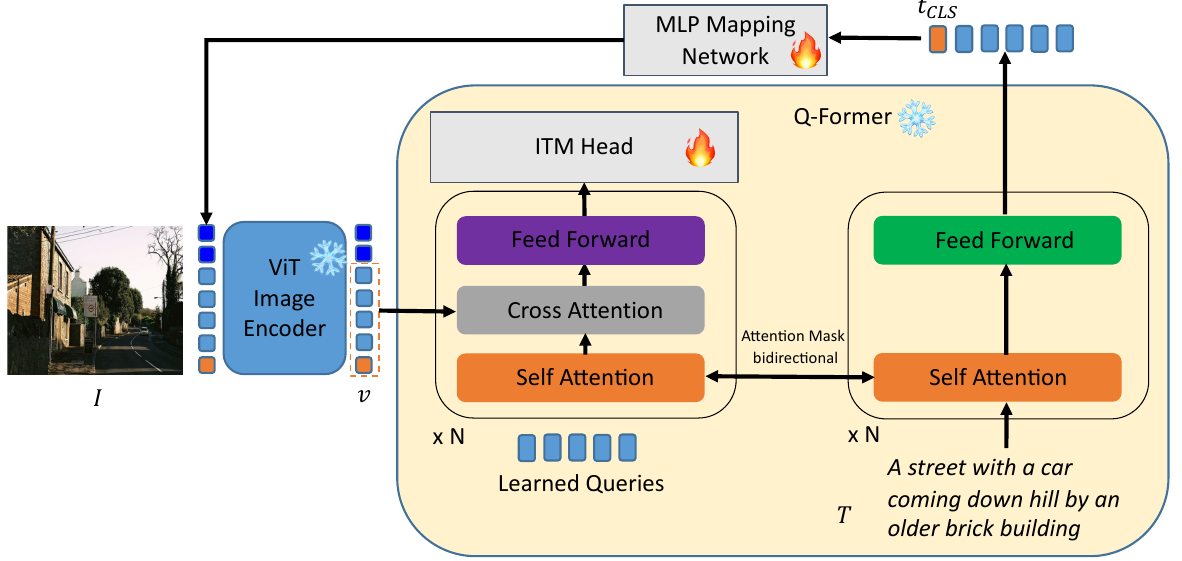}
\vspace{-2mm}
\caption{\textbf{Architecture of ELIP-B.} Similar to the architecture on CLIP/SigLIP, the \emph{MLP Mapping Network} maps the text feature to the visual embedding space. The only difference is the text-guided image features are further fed into the Q-Former to cross-attend the input text and then passed through the Image-Text Matching (ITM) Head to predict whether the image and text match or not. As the input image features to the ITM head have been changed, we also fine-tune the ITM head, which is a lightweight MLP network. The network is fed pairs of text and positive/negative image features at training time and is trained with binary cross entropy loss.}
\vspace{-3mm}
\label{fig:arch-2}
\end{figure}

Here, we propose a mapping network that projects the embedding of the text query into a set of prompt vectors within the visual embedding space. This set of prompt vectors is then incorporated as additional tokens into the first layer of the Vision Transformer (ViT) image encoder, used to re-compute the visual embeddings:
\begin{align*}
&[t_p^1, \ \dots, \ t_p^m, \ t_{\texttt{CLS}}] \ = \ \Phi_{\text{t}}(T)\\
&v \ = \ \Phi_{\text{v}}([x_p^1, \ \dots, \ x_p^n, \ x_{\texttt{CLS}}; \ \psi_{\text{map}}(t_{\texttt{CLS}})])
\end{align*}
where $T$ denotes the query text, which is first encoded with a pre-trained, frozen text encoder~($\Phi_{\text{t}}(\cdot)$) into $m+1$ embeddings. 
The \texttt{[CLS]} token is further fed into a \textbf{trainable} mapping network to generate the prompt vectors, which are concatenated with the $n+1$ image embeddings~($[x_p^1, \dots, x_p^n, x_{\texttt{CLS}}]$), and passed into the pre-trained, frozen visual encoder~($\Phi_{\text{v}}(\cdot)$). 
The MLP Mapping Network consists of 3 layers of linear layers with a GELU between every two linear layers. We expand the output dimension to be $n$ times when we generate $n$ tokens and then divide the generated vector into $n$ tokens. 
The ELIP architecture is shown in Figure~\ref{fig:arch-1} and Figure~\ref{fig:arch-2}.

\subsection{Training and Testing ELIP-C/ELIP-S}
\label{sec:arch-clip}

\noindent \textbf{Text-Guided Contrastive Training.}
At training time, we compute the dot product between the \texttt{[CLS]} token embedding of the text query~($t_{\texttt{CLS}}$) and the re-computed image features guided by the query text, {\em i.e.}, $\{v_1, \dots, v_b\}$~($b$ denotes the batch size). For ELIP-C, we train with the standard InfoNCE loss on the batches; 
For ELIP-S/ELIP-S-2, we train with pairwise Sigmoid loss. 
In Section~\ref{sec:hard_mining}, we provide more details on the batch selection scheme via global hard sample mining.

\vspace{2pt} \noindent \textbf{Re-Ranking at Inference Time.} 
At inference time, for each text query, 
we first compute the similarity scores between the visual-language embedding, computed by the original CLIP/SigLIP model, to obtain an initial ranking of all images. We then select the top-$k$ candidates for further re-ranking, where, the visual features are re-computed by incorporating the prompted vectors from the mapping network. The final ranking is obtained via the dot product of the re-computed image features and the text feature.

\subsection{Training and Testing ELIP-B}
\label{sec:arch-blip}

Figure~\ref{fig:arch-2} illustrates the application of our architecture on BLIP-2. The only difference with that described for CLIP-type models is that BLIP-2 re-ranking does not use a dual encoder, rather, the image and text encoders attend to each other. However, the purpose of our mapping network, and its training, are essentially unchanged.

\vspace{2pt}
\noindent \textbf{Text-Guided Image-Text Matching Loss.}
At training time, we feed the text query ($T$) and the re-computed image features with the query text as prompts, \emph{i.e.,$\{v_+, v_-\}$}~($v_+$ denotes the positive image and $v_-$ denotes the negative image), into the Q-Former, and then to an Image-Text Matching (ITM) Head to predict a score indicating whether the text and image match or not. The output of the ITM head is trained with binary cross entropy loss.

\vspace{2pt}
\noindent \textbf{Inference Time Re-Ranking.}
For each text query, we first compute the similarity scores between the visual-language embedding, computed by the original BLIP-2 image and text encoders, to obtain an initial ranking of all images. We then select the top-$k$ candidates for further re-ranking, where, the visual features are re-computed by incorporating the prompted vectors from the mapping network. The final ranking is obtained via the sum of the initially computed similarity score and the score predicted by the ITM head based on the re-computed image features and text query.

\section{Best Practice for Data Curation \& Training}

The recent visual-language foundation models are often trained on massive numbers (billions) of paired image-caption samples, with considerable computing resources. Here, we explore a `student friendly' \emph{best practice} for data curation that enables improving large-scale visual-language models with limited resources.
Specifically, there are two major challenges to be addressed:
(i) training with a large batch size is challenging, due to limitation on GPU memory; (ii) training on billions of samples is prohibitively expensive on computation cost. 
In Section~\ref{sec:hard_mining}, we discuss the strategy of global hard sample mining to make the training more effective with a small batch size. 
In Section~\ref{sec:large_training_data}, we detail the procedure for selecting and curating an image-text training dataset with maximum information.

\subsection{Global Hard Sample Mining}
\label{sec:hard_mining}

Training CLIP, SigLIP and BLIP-2 often requires a large batch size,
as this increases the chances of having hard training samples, 
and leads to improved contrastive and discriminative capability of the model. Here, we adopt a strategy of global hard sample mining to group hard samples in batches to make training with small batch size more effective. 

\begin{figure}[t]
	\centering
\includegraphics[height=0.5\linewidth]{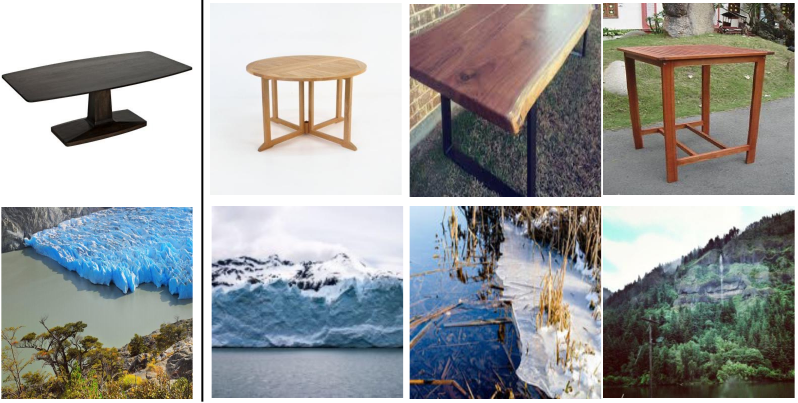}
\vspace{-5mm}
\caption{
\textbf{Examples of generated training batches via global hard sample mining.} 
For each row, the first sample is used to group other samples. 
Captions for row 1~(from left to right): `a wooden table with no base'; `a wooden table with a couple of folding legs on it'; 
`a table that has a metal base with an olive wood top'; 
`small table outdoors sitting on top of the asphalt'. 
Captions for row 2 (from left to right): 
`a huge body of blue ice floats in a mountain stream'; 
`the big chunk of glacier is falling off of the cliff'; 
`there is a broken piece of glass that has been broken from the ground'; `a body of water surrounded by a forest near a mountain'. 
It can be observed that the images and captions are very similar to each other, and significantly more close than images and captions in a random batch. 
} 
\vspace{-3mm}
\label{fig:hard_sample}
\end{figure}

More specifically, for each pair of image-text pair $I_i$ and $T_i$, we compute their image and text features using pre-trained CLIP image and text encoders; 
then we group a batch by collecting other image-text pairs whose images have high CLIP feature similarity scores with the \emph{reference text} $T_i$. 
Examples of our generated training batches are in Figure~\ref{fig:hard_sample}. Assuming the training batch size is $B$ and the original dataset has $N$ image-text pairs, the algorithm gives us $B \times N$ training samples grouped by batches. In practice, we train our model on a random subset of it. 

\subsection{Selection and Curation of Large-Scale Dataset}
\label{sec:large_training_data}

In the literature, a number of large-scale image-text training datasets have been introduced, such as CC3M~\cite{sharma2018conceptual}, DataComp~\cite{gadre2024datacomp}, etc. 
A recent effort~\cite{vasu2024mobileclip} utilises large-scale pre-trained image captioning models to generate synthetic captions for DataComp images, providing more information for training. 
Experiments~\cite{vasu2024mobileclip} show that training CLIP on the generated DataCompDR12M dataset achieves better performance than on DataComp1B, although only 1\% data samples are used. However, in our case, even using DataCompDR12M to train our models still takes prohibitive time (about 200 GPU days) to train ELIP-B on the 12M data with A6000/A40 GPUs.

To accelerate the ELIP-B training, we adopt the batch selection strategy,
that construct batches by learnability, as inspired by JEST~\cite{evans2024data}. Specifically, we run both ELIP-B (the learner) and the pre-trained BLIP-2 model (the reference model) on the grouped batches by our global hard sample mining strategy as in Section~\ref{sec:hard_mining}. We therefore select the batches with the top 10\% highest learnability, where the learnability of a batch is calculated as the difference between the losses of our model and the reference model.

\begin{figure}[t]
	\centering
\includegraphics[height=0.5\linewidth]{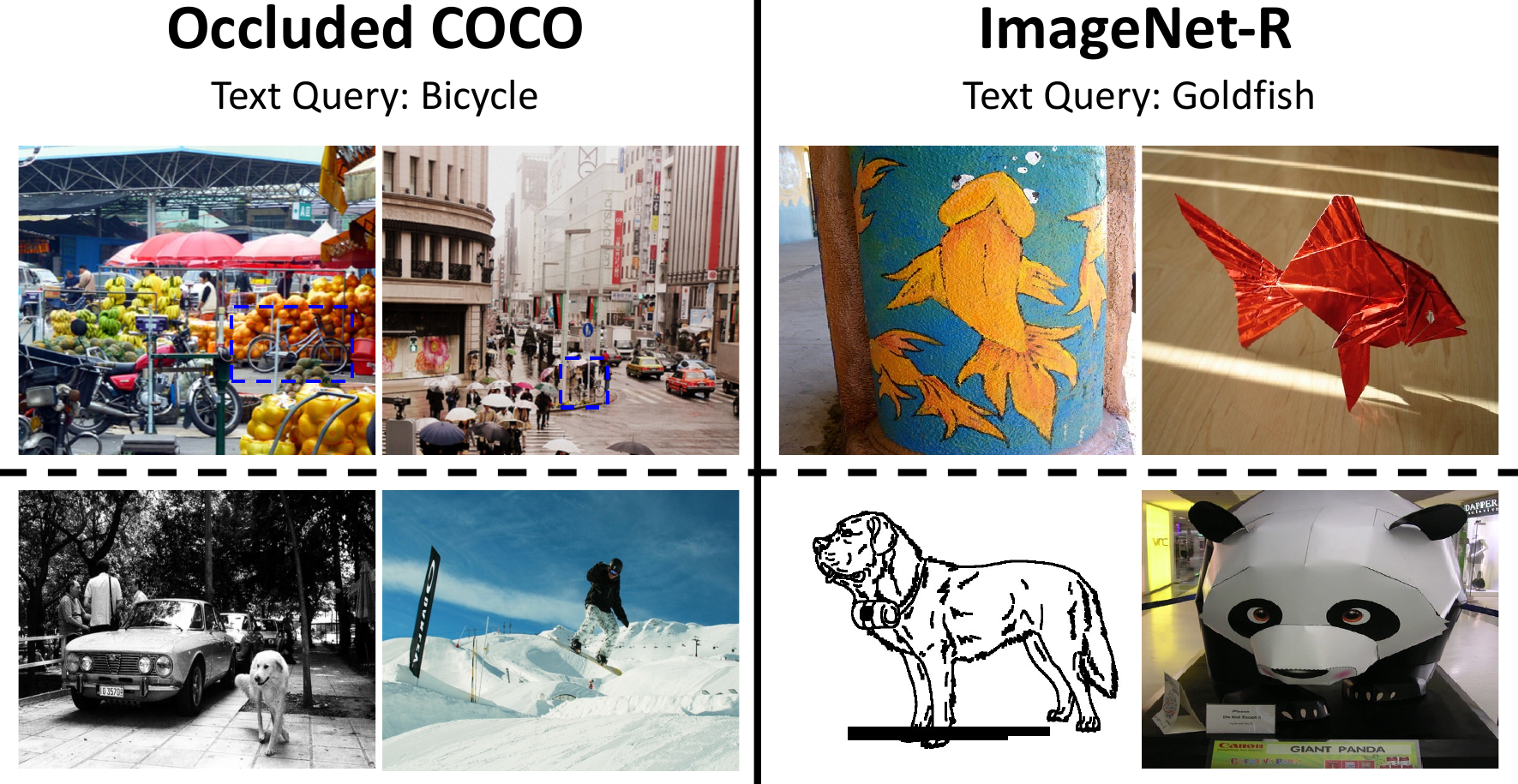}
\vspace{-2mm}
\caption{
\textbf{Examples of the out-of-distribution benchmarks.} Occluded COCO is on the left, and  ImageNet-R is on the right. For both benchmarks, the positive images contain the object described by the text query while the negative images do not contain the object. We display positive images in the first row and negative images in the second row.
For Occluded COCO, the target object in the image is occluded, making it more difficult to be retrieved. For example, for the text query \emph{Bicycle} in Occluded COCO, positive images have an occluded bicycle (highlighted in dashed box) while negative images do not have a bicycle in it; for the text query \emph{Goldfish} in ImageNet-R, positive images have goldfish while negative images do not have goldfish.
} 
\vspace{-3mm}
\label{fig:ood_benchmark}
\end{figure}

\begin{table*}[t]
    \centering
    \tabcolsep=0.10cm
    \begin{tabular}{c|cccc|cccccccc}
    \toprule 

     \multirow{2}{*}{Setting} &\multirow{2}*{\thead{Architecture}} &\multirow{2}*{\thead{Training \\ Dataset}}
     &\multirow{2}*{\thead{Hard Sample \\ Mining}} 
     &\multirow{2}*{\thead{Multiple \\ Prompts}}
     & \multicolumn{3}{c}{COCO} & \multirow{2}*{Avg.} & \multicolumn{3}{c}{Flickr}& \multirow{2}*{Avg.} \\
     \cline{6-8} \cline{10-12}
     & & & & & R@1 & R@5 & R@10 &  & R@1 & R@5 & R@10 & \\
    
    \midrule
    $\mathbb{A}$ & CLIP & - & &  & 40.2 & 66.0 & 75.6 & 60.6 & 67.6 & 88.3 & 93.0 & 83.0 \\ 
    $\mathbb{B}$ & ELIP-C & CC3M~\cite{sharma2018conceptual} & & & 40.7 & 66.2 & 76.1 & 61.0 & 68.8 & 88.9 & 93.8 & 83.8 \\ 
    $\mathbb{C}$ & ELIP-C & CC3M~\cite{sharma2018conceptual} & \checkmark & & 41.8 & 67.5 & 77.5 & 62.3 & 69.5 & 89.7 & 94.1 & 84.4 \\ 
    $\mathbb{D}$ &ELIP-C & DataCompDR~\cite{vasu2024mobileclip} & \checkmark  & & 44.2 & 70.0 & 79.5 & 64.6 & 71.3 & 90.6 & 94.4 & 85.4 \\ 
    $\mathbb{E}$ &ELIP-C & DataCompDR~\cite{vasu2024mobileclip} & \checkmark  & \checkmark & \textbf{45.6} & \textbf{71.1} & \textbf{80.4} & \textbf{65.7} & \textbf{72.3} & \textbf{90.6} & \textbf{94.7} & \textbf{85.9} \\ 
    \bottomrule
    \end{tabular}
    \caption{\textbf{Ablation study on ELIP-C} for choice of training dataset, hard sample mining, and number of prompt vectors generated.
    }
    \label{tab:ablation}
\end{table*}

\begin{table*}[h]
    \centering
    \tabcolsep=0.1cm
    \begin{tabular}{lccccccccc}
    \toprule
 \multirow{2}*{Model} &\multirow{2}*{Year} & \multicolumn{3}{c}{COCO} & \multirow{2}*{Average} & \multicolumn{3}{c}{Flickr}& \multirow{2}*{Average} \\
\cmidrule(lr){3-5} \cmidrule(lr){7-9}
& & Recall@1 & Recall@5 & Recall@10 &  & Recall@1 & Recall@5 & Recall@10 & \\
\midrule 

\emph{CLIP}~\cite{radford2021learning,openclip}& 2021 & 40.16 & 65.95 & 75.62 & 60.58 & 67.56 & 88.34 & 93.00 & 82.97\\ 
    \emph{ELIP-C(Ours)} & - & \textbf{45.61} & \textbf{71.08} & \textbf{80.43} & \textbf{65.71} & \textbf{72.30} & \textbf{90.62} & \textbf{94.68} & \textbf{85.87} \\ 

\midrule
\emph{SigLIP}~\cite{zhai2023siglip}& 2023 & 54.21 & 76.78 & 84.24 & 71.74 & 82.96 & 96.10 & 98.04 & 92.37 \\ 
    \emph{ELIP-S(Ours)} & - & \textbf{61.03} & \textbf{82.62} & \textbf{88.70} & \textbf{77.45} & \textbf{87.62} & \textbf{98.16} & \textbf{99.16} & \textbf{94.98} \\ 

\midrule
\emph{SigLIP-2}~\cite{tschannen2025siglip2}& 2025  & 56.87 & 78.79 & 85.49 & 73.72 & 83.94 & 96.62 & 98.20 & 92.92 \\ 
    \emph{ELIP-S-2(Ours)} & - & \textbf{62.91} & \textbf{83.86} & \textbf{89.70} & \textbf{78.82} & \textbf{87.74} & \textbf{97.96} & \textbf{98.94} & \textbf{94.88} \\ 
    
\midrule
    \emph{BLIP-2*}~\cite{li2023blip}& 2023 & 68.25 & 87.72 & 92.63 & 82.87 & 89.74 & 98.18 & 98.94 & 95.62\\ 
    Q-Pert.(E)*~\cite{sogi2024object} & 2024 & 68.34 & 87.76 & 92.63 & 82.91 & 89.82 & 98.20 & 99.04 & 95.69\\ 
    Q-Pert.(D)*~\cite{sogi2024object} & 2024 & 68.35 & 87.72 & 92.65 & 82.91 & 89.86 & 98.20 & 99.06 & 95.71\\ 
    \emph{ELIP-B(Ours)*} & - & \textbf{68.41} & \textbf{87.88} & \textbf{92.78} & \textbf{83.02}& \textbf{90.08} & \textbf{98.34} & \textbf{99.22} & \textbf{95.88}\\
    \bottomrule
    \end{tabular}
    \caption{
\textbf{Comparison with recent state-of-the-art methods.} Top: CLIP-based models; Middle: SigLIP-based models; Bottom: BLIP-2-based models. 
ELIP-C/ELIP-S brings a significant \textbf{zero-shot} performance boost of CLIP/SigLIP architectures, and ELIP-B outperforms the state-of-the-art BLIP-2 model. 
    Results for models without * are zero-shot, whereas results for models with * are only zero-shot on Flickr as the BLIP-2 model has been fine-tuned on COCO and the * models are based on BLIP-2. 
    However, our method brings an improvement over BLIP-2 on both benchmarks when trained on DataCompDR.
    }
    \label{tab:compare_sota}
\end{table*}

\section{Evaluation Datasets}

We evaluate our models on both the standard text-to-image retrieval benchmarks, COCO~\cite{lin2014microsoft} and Flickr~\cite{plummer2015flickr30k} (Section~\ref{sec:standard_benchmark}), as well as on out-of-distribution benchmarks (Section~\ref{sec:ood_benchmark}) that we newly set up.

\subsection{Standard Benchmarks}
\label{sec:standard_benchmark}

\noindent \textbf{COCO}~\cite{lin2014microsoft},
is a large-scale dataset for studying object detection, segmentation, and captioning. In terms of captioning, each image is annotated with 5 different captions. Previous works use the test split of 5,000 images and 25,010 captions for the evaluation of text-to-image retrieval. 

\vspace{2pt} \noindent \textbf{The Flickr30k Dataset}~\cite{plummer2015flickr30k} contains images collected from Flickr, together with 5 reference sentences provided by human annotators. The test set for text-to-image retrieval consists of 1,000 images and 5,000 captions.

\vspace{2pt} \noindent \textbf{Evaluation Metrics.} 
We adopt the standard metrics for assessing retrieval performance,
namely, Recall@1, Recall@5 and Recall@10. Recall@k denotes the proportion of relevant images that are successfully retrieved within the top-$k$ results for each text query.

\subsection{Out-of-Distribution Benchmarks}
\label{sec:ood_benchmark}

To evaluate a model's capability for text-to-image retrieval in out-of-distribution~(OOD) scenarios, we set up two new benchmarks for text-based image retrieval.
Specifically, \emph{Occluded COCO} focuses on the retrieval of occluded objects, and \emph{ImageNet-R} emphasizes retrieving objects from various unusual domains such as cartoons, sketches, etc. 
Figure~\ref{fig:ood_benchmark} shows examples from the Occluded COCO and ImageNet-R benchmarks.

\vspace{2pt}\noindent \textbf{Occluded COCO} 
is curated with annotations from~\cite{lee2022instance}, 
with the method as described in~\cite{zhan2022tri}, where the occlusion relationship is utilised to collect images containing occluded objects. 
This dataset aims to evaluate the model's performance on retrieving images with occluded target objects against images that do not contain the target object. It has 76 text queries and 5,000 images.

\vspace{2pt} \noindent \textbf{ImageNet-R} is generated using annotations from~\cite{hendrycks2021many} and aims to examine the model's performance for retrieval across various domains, 
for example, art, cartoons, deviantart, graffiti, embroidery, graphics, origami, paintings, patterns, plastic objects, plush objects, sculptures, sketches, tattoos, toys, and video games. It has 200 text queries and 30,000 images.

\vspace{2pt} \noindent \textbf{Evaluation Metrics.} 
Here, we use mAP as the evaluation metric. This is because there might be multiple positive images for each text query.

\section{Experiment}
\label{sec:experiment}

\noindent \textbf{Implementation Details.}
Due to computational resource constraints, 
we train the ELIP-C model with a batch size of 40, 
the ELIP-S model with a batch size of 10, 
and the ELIP-B model with a batch size of 12. 
The initial learning rate is set to $1 \times 10^{-3}$ for ELIP-C, ELIP-S, and ELIP-S-2, and $1 \times 10^{-5}$ for ELIP-B. 
All models are trained on the DataCompDR dataset by default, 
with additional experiments conducted on the smaller CC3M dataset for ablation studies. Training is performed on two A6000 or A40 GPUs. 
For re-ranking, we select the top-$k$ samples based on the dataset and model: for ELIP-C, $k$ is set to 100 for COCO and Flickr, 500 for Occluded COCO, and 1000 for ImageNet-R; for ELIP-S and ELIP-S-2, $k$ is set to 100 for COCO and Flickr, 500 for Occluded COCO, and 200 for ImageNet-R; for ELIP-B, $k$ is set to 20 for COCO and Flickr, 100 for Occluded COCO, and 200 for ImageNet-R. 
The value of $k$ is chosen to ensure high recall in the original ranking while maintaining fast inference. 
Compared to the original pre-training approach of CLIP, SigLIP and BLIP-2, our method significantly improves training efficiency in terms of reduced training time, GPU requirements, and batch size, with only a marginal increase in FLOPS introduced by the trainable MLP mapping network. Further details are provided in the appendix.

\subsection{Results on COCO and Flickr Benchmarks}
\label{sec:results_coco_flickr}

\noindent \textbf{Ablation Study.}
In Table~\ref{tab:ablation}, we evaluate the contributions of different components of the ELIP framework for CLIP. 
A comparison between Settings $\mathbb{A}$ and $\mathbb{B}$ highlights the effectiveness of the ELIP-C boost over the original CLIP. 
The comparison between Settings $\mathbb{B}$ and $\mathbb{C}$ demonstrates the importance of hard sample mining when training with a small batch size. Settings $\mathbb{C}$ and $\mathbb{D}$ show the benefit of training on larger datasets with less noisy captions. Finally, the comparison between Settings $\mathbb{D}$ and $\mathbb{E}$ reveals that generating multiple visual prompts ({\em e.g.}, 10 prompts in this study) is more beneficial than generating a single prompt. Further ablation studies on the number of generated prompts are detailed in the appendix.

\vspace{2pt} \noindent \textbf{Comparison with State-of-the-Art.}
As shown in Table~\ref{tab:compare_sota}, we compare our models~(ELIP-C, ELIP-S, ELIP-S-2, and ELIP-B) with prior state-of-the-art methods. When trained on DataCompDR12M, our method demonstrates zero-shot performance improvements for CLIP, SigLIP, SigLIP-2, and BLIP-2 on the COCO and Flickr benchmarks. Notably, ELIP-B outperforms the most recent work~\cite{sogi2024object}, establishing a new state-of-the-art for text-to-image retrieval on the BLIP-2 backbone. Furthermore, our ELIP-S, when applied to SigLIP and SigLIP-2, achieves performance comparable to BLIP-2. We have also compared ELIP with several baseline methods for re-ranking in the appendix.

\begin{figure}[t]
\centering
\includegraphics[height=0.4\linewidth]{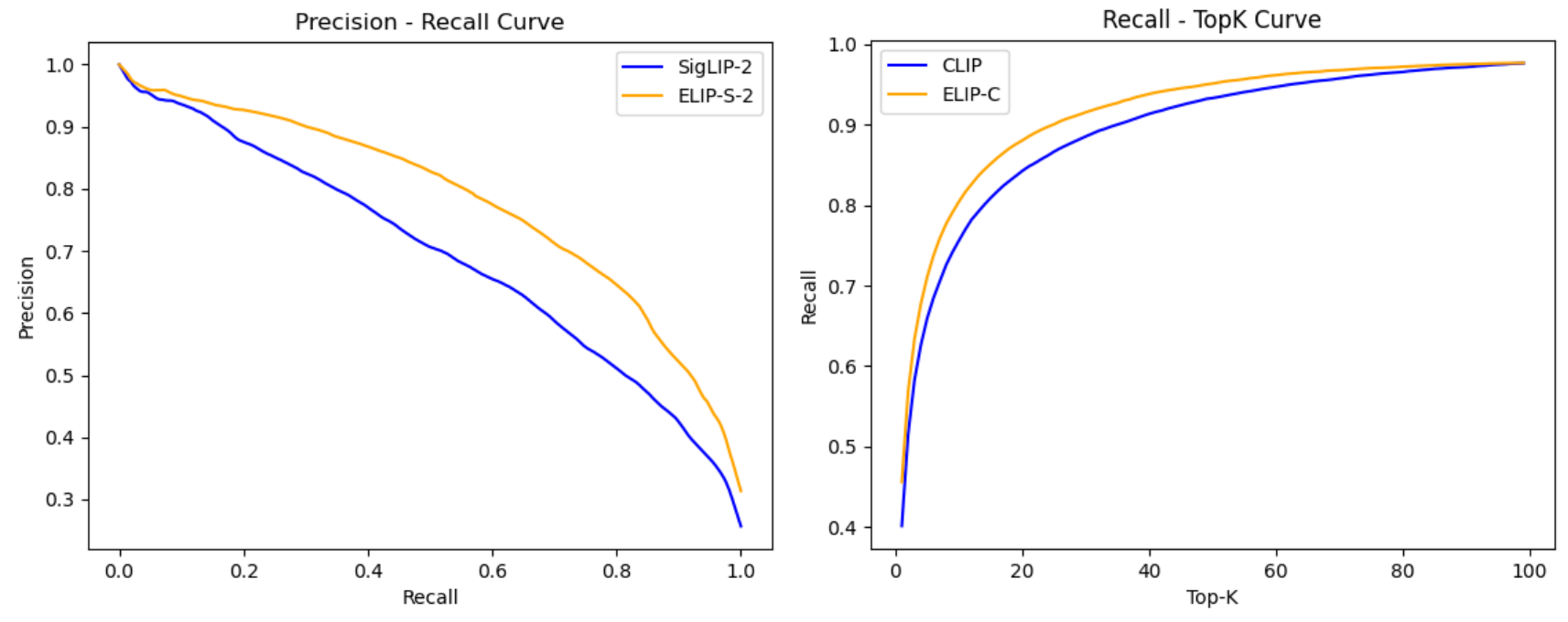}
\vspace{-7mm}
\caption{
\textbf{Before/after comparisons.} Left: Precision-Recall curves for  Occluded COCO retrieval, comparing SigLIP-2 initial rankings to the re-rankings given by ELIP-S-2. Right:  Recall Top-$k$ curves for COCO retrieval, comparing  CLIP initial rankings to the re-rankings given by ELIP-C. 
} 
\label{fig:curve_recall}
\end{figure}

\vspace{2pt}
\noindent \textbf{Recall Top-$k$ Curves.}
Figure~\ref{fig:curve_recall} (right) presents the Recall@Top-$k$ curves for the original CLIP model and our ELIP-C on the COCO benchmark. The curves are generated by plotting the Recall values across various Top-$k$ thresholds. Notably, there is a significant performance gap between the two models, demonstrating that ELIP-C re-ranking consistently improves text-to-image retrieval performance across different $k$ values.

\begin{figure}[t]
\centering
\includegraphics[height=1.35\linewidth]{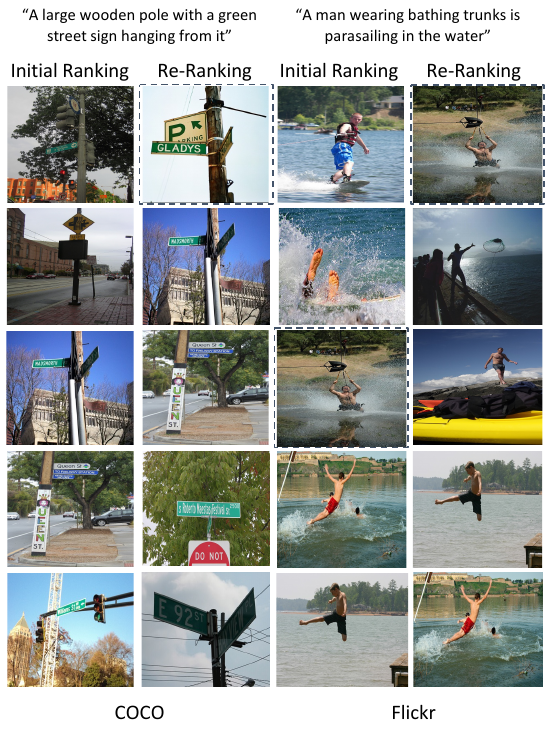}
\vspace{-4mm}
\caption{
\textbf{Qualitative comparison between CLIP initial ranking and ELIP-C re-ranking.} COCO: Columns 1–2; Flickr: Columns 3–4. 
The ground-truth image for each query is highlighted with a dashed box, with the top-5 retrieved images shown. For the COCO query ``A large wooden pole with a green street sign hanging from it'', CLIP ranks a non-wooden pole as top-1, while ELIP-C correctly re-ranks the large wooden pole to top-1. For the Flickr query ``A man wearing bathing trunks is parasailing in the water'', CLIP ranks a wakeboarding person as top-1, whereas ELIP-C accurately re-ranks the parasailing man wearing bathing trunks to top-1.} 
\label{fig:qualitative}
\vspace{-4mm}
\end{figure}

\vspace{2pt}
\noindent \textbf{Qualitative Results.}
Figure~\ref{fig:qualitative} provides a qualitative comparison between the initial rankings produced by the CLIP model and the re-ranked results obtained with ELIP-C on the COCO~(left) and Flickr~(right) benchmarks. In both cases, ELIP-C significantly improves the rankings by elevating the ground-truth image (highlighted with a dashed box) to rank 1. Additional qualitative results are provided in the appendix.

\begin{figure*}[h]
\centering
\includegraphics[height=0.3\linewidth]{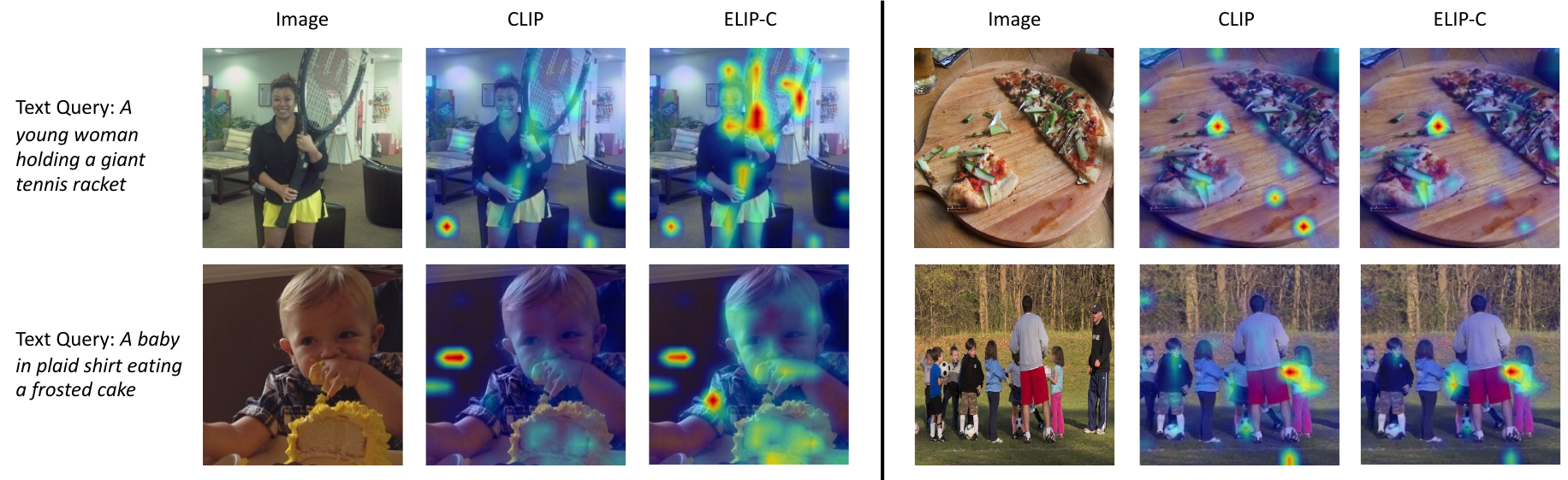}
\vspace{-2pt}
\caption{
\textbf{Visualisation of attention maps} comparing the cross-attention maps of the \texttt{[CLS]} token on patch tokens for CLIP and ELIP-C. Left: image matches the text query; Right: image does not match the text query. For matched queries, ELIP-C enhances attention on image features relevant to the text. 
For example: (Row 1) ELIP-C focuses more on the giant tennis racket and the young woman for the query ``A young woman holding a giant tennis racket''; 
(Row 2) ELIP-C highlights the cake, baby, and shirt for the query ``A baby in plaid shirt eating a frosted cake''.
Differences are minimal when the image does not match the query (Columns 4–6). 
} 
\label{fig:attention_map}
\end{figure*}

\vspace{2pt} \noindent \textbf{Visualisation of Attention Map.} 
Figure~\ref{fig:attention_map} visualizes the cross-attention maps of the \texttt{[CLS]} token on patch tokens for both CLIP and ELIP-C on COCO. When the image matches the text query (left of Figure~\ref{fig:attention_map}), our generated visual prompt vectors effectively enhance the selection of image features relevant to the query. This improvement can be attributed to ELIP-C's early fusion approach, which integrates text features at the beginning of the image encoder, enabling the model to produce image embeddings more closely aligned with the query text. The visualizations provide strong evidence supporting this hypothesis.

\begin{table}[h]
    \centering
    \tabcolsep=0.15cm
    \begin{tabular}{c|ccc}
    \toprule 
    Model & \thead{Occluded \\ COCO} & ImageNet-R & Average  \\ 
    \midrule
    CLIP & 47.47 & 76.01 & 61.74 \\ 
    ELIP-C~(zero-shot) & 48.89 & 76.81 & 62.85 \\ 
    ELIP-C~(fine-tuned) & \textbf{59.88} & \textbf{81.44} & \textbf{70.66} \\

    \midrule
    SigLIP & 61.74 & 92.11 & 76.93 \\ 
    ELIP-S~(zero-shot) & 64.58 & 92.42 & 78.50 \\ 
    ELIP-S~(fine-tuned) & \textbf{71.99} & \textbf{92.86} & \textbf{82.43} \\ 
    
    \midrule
    SigLIP-2 & 66.40 & 92.66 & 79.53 \\ 
    ELIP-S-2~(zero-shot) & 67.42 & 92.74 & 80.08 \\ 
    ELIP-S-2~(fine-tuned) & \textbf{76.10} & \textbf{94.00} & \textbf{85.05} \\ 

    \midrule
    BLIP-2 & 62.73 & 82.31 & 72.52 \\ 
    ELIP-B~(zero-shot) &  63.40 & 82.99 & 73.20 \\ 
    ELIP-B~(fine-tuned) & \textbf{70.49} & \textbf{83.68} & \textbf{77.09} \\ 
    
    \bottomrule
    \end{tabular}
    \caption{\textbf{mAP Results on OOD datasets.} 
    ELIP prompting achieves notable zero-shot improvements for CLIP, SigLIP series, and BLIP-2. These gains are further amplified through fine-tuning on relevant datasets. 
    For example, to adapt to the Occluded COCO, the ELIP model is fine-tuned on COCO. Similarly, fine-tuning on ImageNet adapts ELIP to ImageNet-R. These results demonstrate ELIP's capability for efficiently adapting the models to new datasets. 
    }
    \label{tab:ood_result}
\end{table}

\subsection{Results on OOD Benchmarks}
\label{sec:ood_result}

The results on the out-of-distribution (OOD) benchmarks are presented in Table~\ref{tab:ood_result}. ELIP achieves notable {\em zero-shot} improvements across all models on the OOD benchmarks, Occluded COCO and ImageNet-R, highlighting the strong generalization capabilities of the ELIP models. 
The performance can be improved further by fine-tuning the mapping network on suitable datasets (the image and text encoders are frozen).
Since it is not feasible to fine-tune on Occluded COCO~(very few data samples) and ImageNet-R~(evaluation only),
for Occluded COCO retrieval we fine-tune on the original COCO dataset, and for ImageNet-R retrieval we fine-tune on ImageNet. As can be seen in Table~\ref{tab:ood_result}  by this fine-tuning the performance of all the models is significantly boosted further. This demonstrates that fine-tuning ELIP enables efficient adaptation of the models to new datasets. 
The significant difference ELIP makes is also illustrated in Figure~\ref{fig:curve_recall} (left). 
Please refer to the appendix on the fine-tuning.

\section{Conclusion}
\label{sec:conclusion}

In this paper, we introduced {\em Enhance Language-Image Pre-training (ELIP)}, a method to improve visual-language foundation models for text-based image retrieval. ELIP is a simple plug-and-play modification to pre-trained visual-language foundation models that significantly improves their zero-shot performance. Furthermore, the mapping network can be fine-tuned to
efficiently adapt these models to OOD datasets, leading to still further improvements. We have also demonstrated, by visualizing the attention maps, that ELIP enables the image encoder to attend to more relevant details.

Future work could apply ideas similar to ELIP to enhance generative Multimodal Large Language Models by introducing more effective text-guided visual attention and encoding for both decoder-only architectures~\cite{liu2023visual} and cross-attention-based architectures~\cite{alayrac2022flamingo}.

\paragraph{Acknowledgements. } 
This research is supported by EPSRC Programme Grant VisualAI EP$\slash$T028572$\slash$1, a Royal Society
Research Professorship RP$\backslash$R1$\backslash$191132, a China Oxford Scholarship and the Hong Kong Research Grants Council -- General
Research Fund (Grant No.: 17211024).
We thank Minghao Chen, Jindong Gu, Zhongrui Gui, Zhenqi He, João Henriques, Zeren Jiang, Zihang Lai, Horace Lee, Kun-Yu Lin, Xianzheng Ma, Christian Rupprecht, Ashish Thandavan, Jianyuan Wang, Kaiyan Zhang, Chuanxia Zheng and Liang Zheng for their help and support for the project.

{
    \small
    \bibliographystyle{ieeenat_fullname}
    \bibliography{main}
}

\onecolumn

\appendix
\section*{Appendix}

\section{Additional Implementation Details}
\label{sec:sup_implement_detail}

In this section, we provide additional implementation details regarding: the choice of $k$ for re-ranking in Section~\ref{sec:supple_choice_of_k} as mentioned in Section~7; the general experiments in Section~\ref{sec:sup_implement_detail_general} as mentioned in Section~5.1 and Section~7; fine-tuning experiments in Section~\ref{sec:sup_implement_detail_finetune} as mentioned in Section~7.2; and the Occluded COCO benchmark in Section~\ref{sec:sup_implement_detail_occluded} as mentioned in Section~6.2.

\subsection{Choice of $k$ for Re-Ranking}
\label{sec:supple_choice_of_k}

\begin{table*}[h]
    \centering
    \tabcolsep=0.25cm
    \begin{tabular}{cccccc}

\toprule
 CLIP & & & & & \\
\midrule 
Dataset & Top-20 & Top-50 & Top-100 & Top-200 & Top-500 \\

COCO & 84.25 & 93.30 & 97.67 & 99.12 & 99.65 \\
Flickr & 95.90 & 98.32 & 99.24 & 99.82 & 99.98 \\

\midrule

Dataset & Top-100 & Top-200 & Top-500 & Top-1000 & Top-2000 \\

Occluded COCO & 65.31 & 72.00 & 81.14 & 87.66 & 93.71 \\
ImageNet-R & 57.33 & 77.83 & 91.13 & 95.05 & 97.18 \\

\bottomrule

\toprule
 SigLIP & & & & & \\
\midrule 
Dataset & Top-20 & Top-50 & Top-100 & Top-200 & Top-500 \\

COCO & 90.14 & 96.07 & 98.77 & 99.44 & 99.78 \\
Flickr & 98.98 & 99.64 & 99.82 & 99.92 & 100.00 \\

\midrule

Dataset & Top-50 & Top-100 & Top-200 & Top-500 & Top-1000 \\

Occluded COCO & 68.84 & 75.90 & 81.18 & 86.96 & 91.94 \\
ImageNet-R & 39.22 & 67.46 & 90.46 & 98.14 & 99.04 \\

\bottomrule

\toprule
 SigLIP-2 & & & & & \\
\midrule 
Dataset & Top-20 & Top-50 & Top-100 & Top-200 & Top-500 \\

COCO & 91.10 & 96.59 & 98.85 & 99.55 & 99.80 \\
Flickr & 99.34 & 99.80 & 99.92 & 99.94 & 100.00 \\

\midrule

Dataset & Top-50 & Top-100 & Top-200 & Top-500 & Top-1000 \\

Occluded COCO & 70.69 & 79.57 & 85.67 & 91.28 & 95.34 \\
ImageNet-R & 39.39 & 67.74 & 90.90 & 98.24 & 99.10 \\

\bottomrule

\toprule
 BLIP-2 & & & & & \\
\midrule 
Dataset & Top-5 & Top-10 & Top-20 & Top-50 & Top-100 \\

COCO & 86.08 & 91.85 & 95.96 & 98.80 & 99.63 \\
Flickr & 97.64 & 99.06 & 99.52 & 99.82 & 99.86 \\

\midrule

Dataset & Top-50 & Top-100 & Top-200 & Top-500 & Top-1000 \\

Occluded COCO & 68.67 & 77.40 & 82.12 & 88.17 & 92.67 \\
ImageNet-R & 35.97 & 61.21 & 82.81 & 92.88 & 94.99 \\

    \bottomrule

    \end{tabular}
    \caption{\textbf{Recall @ Different $k$ for CLIP, SigLIP and BLIP-2 initial ranking.} See text for a more detailed discussion.}
    \label{tab:choice_k}
\end{table*}

In terms of the selection of the top-$k$ values for re-ranking as mentioned in Section~7, the value of $k$ is chosen such that the recall at that $k$ is high in the original ranking, and also so that the inference is fast. 
For ELIP-C, $k$ is set to 100 for COCO and Flickr, 500 for Occluded COCO, and 1000 for ImageNet-R; for ELIP-S and ELIP-S-2, $k$ is set to 100 for COCO and Flickr, 500 for Occluded COCO, and 200 for ImageNet-R; for ELIP-B, $k$ is set to 20 for COCO and Flickr, 100 for Occluded COCO, and 200 for ImageNet-R. 
More specifically, as in Table~\ref{tab:choice_k}, for 
CLIP, the recall@100 for COCO/Flickr is over 95\%, recall@500 for Occluded COCO is over 80\%, recall@1000 for ImageNet-R is over 95\%; For SigLIP/SigLIP-2, the recall@100 for COCO/Flickr is over 95\%, recall@500 for Occluded COCO is over 85\%, recall@200 for ImageNet-R is over 90\%; for BLIP-2, the recall@20 for COCO/Flickr is over 95\%, recall@100 for Occluded COCO is over 75\%, recall@200 for ImageNet-R is over 80\%. 
We have experimented with increasing $k$ for re-ranking, but there is no significant performance boost while the computational cost is much higher.

\subsection{Implementation Details for General Experiments}
\label{sec:sup_implement_detail_general}

As mentioned in Section~5.1, in practice we use a random subset of 6M samples for training ELIP-C/ELIP-S and a random subset of 1M samples for training ELIP-B. For ELIP-B, we first select the samples with the top 10\% highest learnability as described in Section~5.2, and then train the ELIP-B architecture with the selected training samples. For ELIP-C, we use the default backbone (ViT-B) given limited computing resources and for the ablations, while for ELIP-S and ELIP-B we use the best backbone (ViT-G for BLIP-2 and SigLIP-2, ViT-SO400M for SigLIP) to show the effectiveness of our method over the state-of-the-art model. 
As we use large backbones for ELIP-S and it increases GPU memory, for every training batch we take 10 samples out of the batch of 40 in the ELIP-C training dataset.

The implementation of the MLP Mapping Network is three linear layers with a GELU between each pair of layers. We expand the output dimension to be $n$ times when we generate $n$ tokens and then divide the generated vector into $n$ tokens. 
We have also investigated other architectures for the Mapping Network, such as transformer layers, different numbers of layers of MLP, and linear network, but there is no significant performance advantage found. We therefore use the current architecture which is simple but effective.

\subsection{Implementation Details for Fine-Tuning on COCO and ImageNet}
\label{sec:sup_implement_detail_finetune}

As mentioned in Section~7.2, we have further fine-tuned our models (pre-trained on DataCompDR) on COCO and ImageNet, respectively. 
For both datasets, we first group the hard samples in batches using the method described in Section~5.1. 
The captions for the images are category names, 
and we ensure that in a batch one category can only appear once, 
so that there will not be noise in training. 
We fine-tune ELIP-C, ELIP-S, and ELIP-B with an initial learning rate of $1e-5$ for 10k iterations. 
Although there is still a domain gap between COCO and Occluded COCO, ImageNet, and ImageNet-R, we find that even fine-tuning on COCO and ImageNet can already bring a boost to the performance on Occluded COCO and ImageNet-R as the caption domain is closer. It is not feasible to fine-tune on Occluded COCO and ImageNet-R because Occluded COCO has very few data samples and ImageNet-R is designed for evaluation.

\subsection{Implementation Details for Setting up the Occluded COCO Benchmark}
\label{sec:sup_implement_detail_occluded}

As mentioned in Section~6.2, we set up the Occluded COCO benchmark using the annotations from~\cite{lee2022instance}. More specifically, for each COCO category, we collect two lists - one for the images containing occluded object of such category, and the other for the images not having object of such category.
For images containing occluded object of such category, 
we select the ones that meet the following criteria: 
1) The image contains object of the category; 2) All instances of such category are occluded by at least one other instance, \emph{i.e.,} have an occludee-occluder relationship with another instance in the annotations of~\cite{lee2022instance}. The idea of collecting images containing occluded instances is similar to~\cite{zhan2022tri}, which utilises the occlusion relationship.

\clearpage
\section{Model Variations and Comparison with Other Re-Ranking Methods}
\label{sec:additional_baseline}

\begin{table}[h]
   \centering
   
   \tabcolsep=0.30cm
   \begin{tabular}{lcccc}
   \toprule
   Model &  ID & Recall@1 & Recall@5 & Recall@10 \\
   \midrule 
   CLIP & - & 67.56 & 88.34 & 93.00 \\
   \midrule 
   Late Fusion Baseline& 1 & 69.58 & 89.58 & 93.66 \\
   Dense Text Token & 2 & 71.66 & 90.44 & 94.54 \\
   Cross-Attention Pooling& 3 & 67.78 & 88.42 & 93.52 \\
   \midrule
   Text Query Expansion& 4 & 66.28 & 87.60 & 92.88 \\
   Image Crop Features& 5 & 67.52 & 88.36 & 92.98 \\
   QE + LF & 6 & 67.18 & 88.44 & 93.00 \\
   \midrule 
   ELIP-C (ours) & - & \textbf{72.30} & \textbf{90.62} & \textbf{94.68} \\
   \bottomrule
   \end{tabular}
   \caption{Model variations for ELIP-C (ID 1-3), and comparison with other re-ranking methods (ID 4-6). The performance is measured on Flickr retrieval. See the text for a detailed description of each method. 
   }
   \vspace{0.4cm}
   
   \label{tab:additional_baseline}
\end{table}

In this section, we investigate variations of the ELIP model and also compare with several other re-ranking methods as mentioned in Section~VI of the main paper.
We report results using CLIP on Flickr. 

\noindent \textbf{Variations on the ELIP model}
\begin{itemize}

\item \textbf{1.\ Late Fusion Baseline:} For both training and inference, the MLP Mapping Network is used to map the text embedding to generate multiple prompt vectors that are inserted into the last layer of the ViT image encoder rather than the first layer.

\item \textbf{2.\ Dense Text Token:} Rather than use the $t_{CLS}$ embedding to generate all prompt vectors, for both training and inference, we use the average pooling of dense text token representations to generate the prompt vectors.

\item \textbf{3.\ Cross-Attention Pooling:} Given the image patch embeddings at the final ViT layer, for both training and inference, we map the $t_{CLS}$ vector to the same dimension as the image patch tokens via a MLP network, and use it to cross-attend to the image patch embeddings to get the image feature, instead of generating prompt vectors and inserting into ViT image encoder. The idea of cross-attention pooling is similar to~\cite{bardes2024revisiting}.

\end{itemize}

\noindent \textbf{Comparisons with other methods}

\begin{itemize}

\item \textbf{4.\ Text Query Expansion:} For each text query, we ask GPT-4~\cite{achiam2023gpt} to rephrase it into 5 variants. At the re-ranking stage, we compute the similarity scores of the top-$k$ images with the 5 variants of text queries, and then take an average of the 5 scores together with the initial score to generate the final score.

\item \textbf{5.\ Image Crop Features:} For each of the top-$k$ images, we randomly crop and resize it into 5 variants and feed them into the CLIP image encoder to get the image local features. Then we compute the similarity scores between the image local features and the text features, and take an average of the 5 scores together with the initial score to generate the final score.

\item \textbf{6.\ Query Expansion + Local Features:} We use the top-3 retrieved images as expanded query~\cite{chum2007total} for re-ranking, \emph{i.e.,} the top-3 retrieved image features are used as query features in the re-ranking.
We then compare the Chamfer Similarity~\cite{barrow1977parametric,kordopatis2025ilias} between patch features of the expanded query images and other top-$k$ images. The patch features are the patch tokens of the ViT image encoder. The average of the Chamfer similarities and the initial score is used as the final score for re-ranking.
We have also tried to directly match image patch tokens with text dense tokens, but the performance is not good due to the fact that text tokens and image patch tokens are not properly aligned.

\end{itemize}

It can be observed that the ELIP-C model achieves the best performance over the variations and in comparison to other methods.

\clearpage
\section{Additional Qualitative Results}
\label{sec:sup_qualitative}

In this section, we provide more qualitative results as mentioned in Section~7 of the main paper, including: 
1) Qualitative comparison between the rankings of CLIP and ELIP-C on COCO (Figure~\ref{fig:supple_qualitative_clip1}), Flickr (Figure~\ref{fig:supple_qualitative_clip1-2}), Occluded COCO (Figure~\ref{fig:supple_qualitative_clip2}) and ImageNet-R (Figure~\ref{fig:supple_qualitative_clip2-2}); 
2) Qualitative comparison between the rankings of SigLIP/SigLIP-2 and ELIP-S/ELIP-S-2 on COCO (Figure~\ref{fig:supple_qualitative_siglip1}), Flickr (Figure~\ref{fig:supple_qualitative_siglip1-2}), Occluded COCO (Figure~\ref{fig:supple_qualitative_siglip2}) and ImageNet-R (Figure~\ref{fig:supple_qualitative_siglip2-2}); 
3) Qualitative comparison between the rankings of BLIP-2 and ELIP-B on COCO (Figure~\ref{fig:supple_qualitative_blip1}), Flickr (Figure~\ref{fig:supple_qualitative_blip1-2}), Occluded COCO (Figure~\ref{fig:supple_qualitative_blip2}) and ImageNet-R (Figure~\ref{fig:supple_qualitative_blip2-2}); 
4) Attention maps for CLIP and ELIP-C (Figure~\ref{fig:supple_attention_clip} and Figure~\ref{fig:supple_attention_clip-2}); 
5) Attention maps for SigLIP/SigLIP-2 and ELIP-S/ELIP-S-2(Figure~\ref{fig:supple_attention_siglip} and Figure~\ref{fig:supple_attention_siglip-2}); 
6) Attention maps for BLIP-2 and ELIP-B (Figure~\ref{fig:supple_attention_blip} and Figure~\ref{fig:supple_attention_blip-2}).

\clearpage
\subsection{Qualitative Comparison of CLIP and ELIP-C}

\begin{figure*}[h]
	\centering
\includegraphics[height=0.75\linewidth]{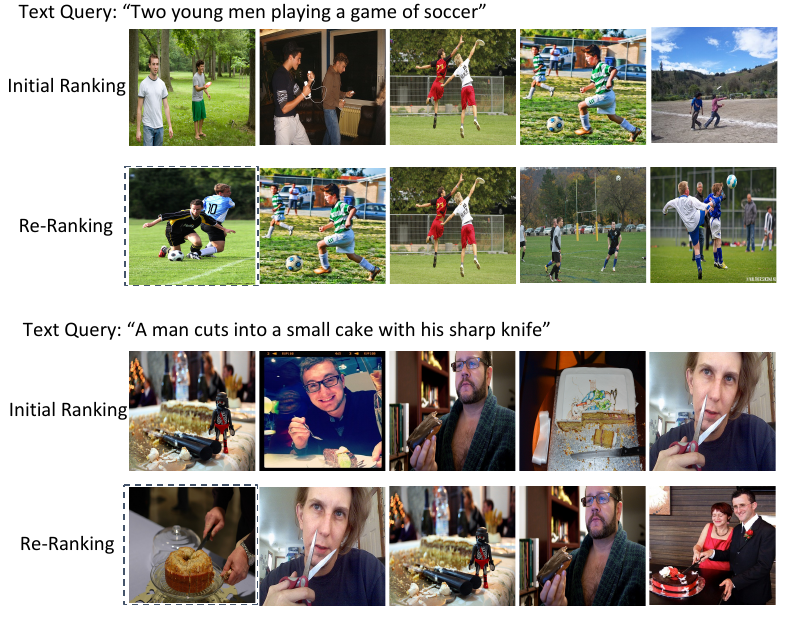}
	\caption{
\textbf{Qualitative Comparison on COCO for CLIP and ELIP-C.} 
For each example, we show both top-5 rankings (from left to right) and highlight the ground truth image in a black dashed box.
In the first and second examples (from top to bottom), the text queries are `Two young men playing a game of soccer', and `A man cuts into a small cake with his sharp knife', and the ground truth images are not in top-5 images of the CLIP initial ranking, but ranked top-1 in our ELIP-C re-ranking. 
	} 
	\label{fig:supple_qualitative_clip1}
	\end{figure*}

Figure~\ref{fig:supple_qualitative_clip1} shows more qualitative comparison between the CLIP initial ranking and ELIP-C re-ranking on COCO. 

\begin{figure*}[h]
	\centering
\includegraphics[height=0.75\linewidth]{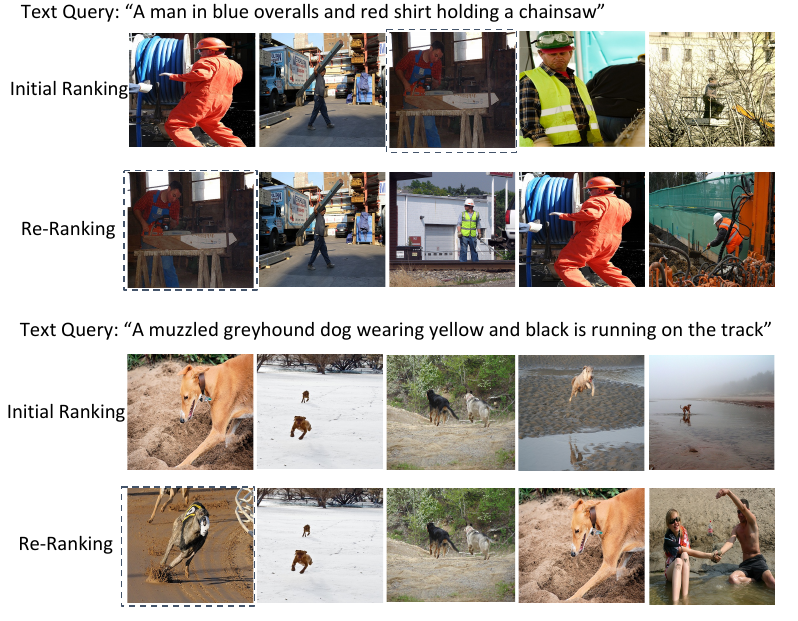}
	\caption{
\textbf{Qualitative Comparison on Flickr for CLIP and ELIP-C.} For each example, we show both top-5 rankings (from left to right) and highlight the ground truth image in a black dashed box.
In the first example, the text query is `A man in blue overalls and red shirt holding a chainsaw', and the ground truth image ranks top-3 in the initial ranking but top-1 in our re-ranking. In the second example, the text query is `A muzzled greyhound dog wearing yellow and black is running on the track', and the ground truth image is not in top-5 images of the CLIP initial ranking, but ranked top-1 in our ELIP-C re-ranking. 
	} 
	\label{fig:supple_qualitative_clip1-2}
	\end{figure*}

\clearpage
Figure~\ref{fig:supple_qualitative_clip1-2} shows more qualitative comparison between the CLIP initial ranking and ELIP-C re-ranking on Flickr.

\begin{figure*}[h]
	\centering
\includegraphics[height=0.75\linewidth]{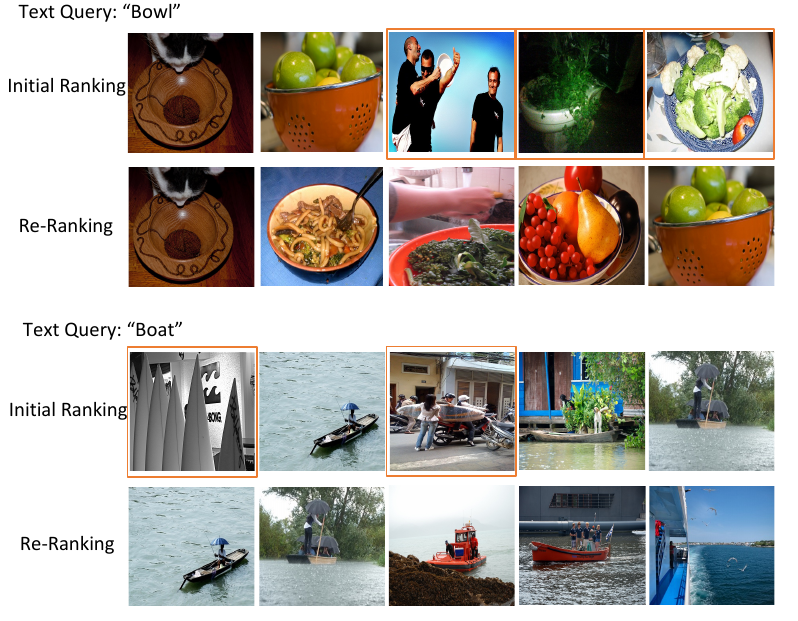}
	\caption{
\textbf{Qualitative Comparison on Occluded COCO for CLIP and ELIP-C.} We display the top-5 retrieved images. Negative samples (errors) are highlighted in an orange solid box. 
For the first example, the text query is `bowl' and CLIP confuses it with `frisbee', `toilet', and `plate' in top-5 retrieved images; for the second example, the text query is `boat' and CLIP confuses it with `surfboard' in top-5 retrieved images.  
	} 
	\label{fig:supple_qualitative_clip2}
	\end{figure*}

\clearpage
Figure~\ref{fig:supple_qualitative_clip2} displays qualitative comparison of CLIP initial ranking and our ELIP-C re-ranking on Occluded COCO. 

\begin{figure*}[h]
	\centering
\includegraphics[height=0.75\linewidth]{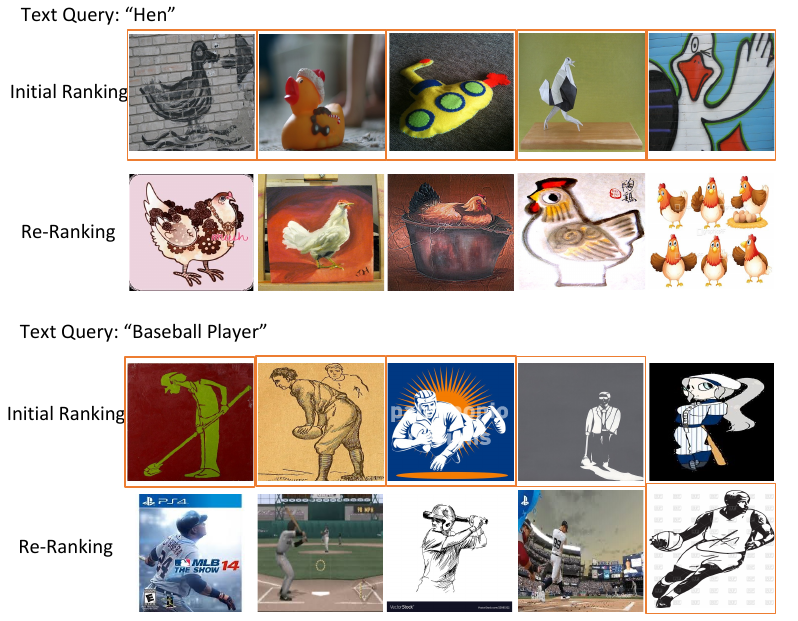}
	\caption{
\textbf{Qualitative Comparison on ImageNet-R for CLIP and ELIP-C.} We display examples at the top-100 rankings where there is a difference between the two models, \emph{i.e.,} one model retrieves a positive sample while the other model retrieves a negative sample, as the majority of the top-100 retrieves samples are positive. \emph{Generally, ELIP-C retrieves more positive samples in top-100 images than CLIP}. Negative samples (errors) are highlighted in an orange solid box. 
For the first example, the text query is `hen' and CLIP retrieves some `duck's in top-100 retrieved images; for the second example, the text query is `baseball player' and CLIP retrieves some players for other ball games in the top-100 retrieved images while our ELIP-C retrieves a basketball player in top-100 images.  
	} 
	\label{fig:supple_qualitative_clip2-2}
	\end{figure*}

\clearpage
Figure~\ref{fig:supple_qualitative_clip2-2} displays qualitative comparison of CLIP initial ranking and our ELIP-C re-ranking on ImageNet-R. 

\clearpage
In conclusion, these results demonstrate that, though the initial CLIP model can roughly retrieve relevant images, it is less able to distinguish fine-grained differences between the top retrieved images, whereas our re-ranking model further improves the discrimination between the ground truth and hard negative images, achieving a better retrieval.

\clearpage
\subsection{Qualitative Comparison of SigLIP/SigLIP-2 and ELIP-S/ELIP-S-2}

\begin{figure*}[h]
	\centering
\includegraphics[height=0.75\linewidth]{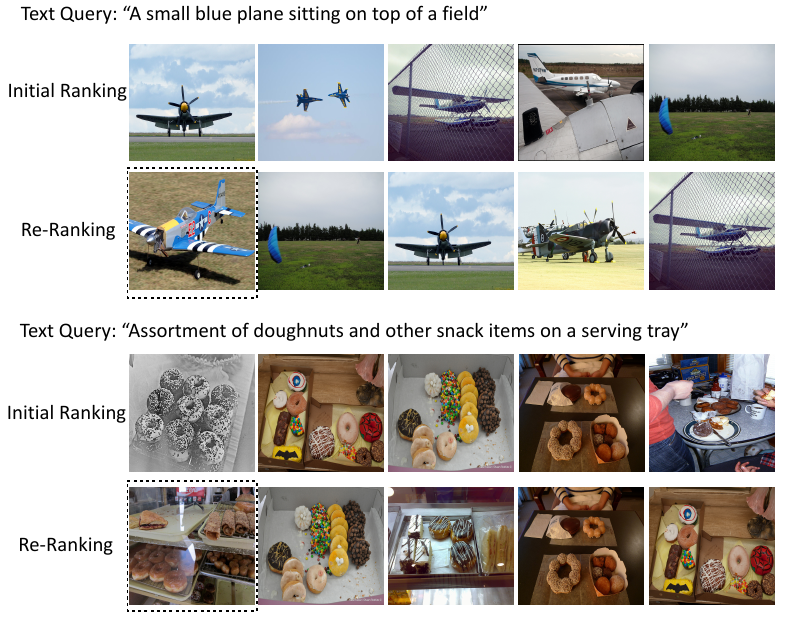}
	\caption{
\textbf{Qualitative Comparison on COCO for SigLIP and ELIP-S (top), SigLIP-2 and ELIP-S-2 (bottom).} For each example, we show both top-5 rankings (from left to right) and highlight the ground truth image in a black dashed box.
In the first (SigLIP v.s. ELIP-S) and second (SigLIP-2 v.s. ELIP-S-2) examples (from top to bottom), the text queries are `A small blue plane sitting on top of a field', and `Assortment of doughnuts and other snack items on a serving tray', and the ground truth images are not in top-5 images of the SigLIP/SigLIP-2 initial ranking, but ranked top-1 in our ELIP-S/ELIP-S-2 re-ranking. 
	} 
	\label{fig:supple_qualitative_siglip1}
	\end{figure*}

Figure~\ref{fig:supple_qualitative_siglip1} shows more qualitative comparison between the SigLIP/SigLIP-2 initial ranking and ELIP-S/ELIP-S-2 re-ranking on COCO. 

\begin{figure*}[h]
	\centering
\includegraphics[height=0.75\linewidth]{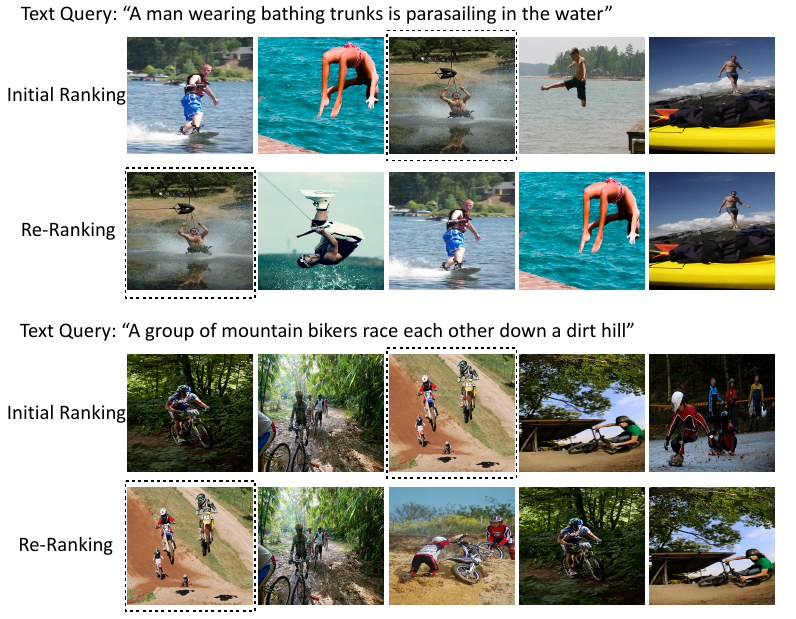}
	\caption{
\textbf{Qualitative Comparison on Flickr for SigLIP and ELIP-S (top), SigLIP-2 and ELIP-S-2 (bottom).} For each example, we show both top-5 rankings (from left to right) and highlight the ground truth image in a black dashed box.
In the first example, the text query is `A man wearing bathing trunks is parasailing in the water', and the ground truth image ranks top-3 in the SigLIP initial ranking but top-1 in our ELIP-S re-ranking. In the second example, the text query is `A group of mountain bikers race each other down a dirt hill', and the ground truth image ranks top-3 in the SigLIP-2 initial ranking, but ranked top-1 in our ELIP-S-2 re-ranking. 
	} 
	\label{fig:supple_qualitative_siglip1-2}
	\end{figure*}

\clearpage
Figure~\ref{fig:supple_qualitative_siglip1-2} shows more qualitative comparison between the SigLIP/SigLIP-2 initial ranking and ELIP-S/ELIP-S-2 re-ranking on Flickr.

\begin{figure*}[h]
	\centering
\includegraphics[height=0.75\linewidth]{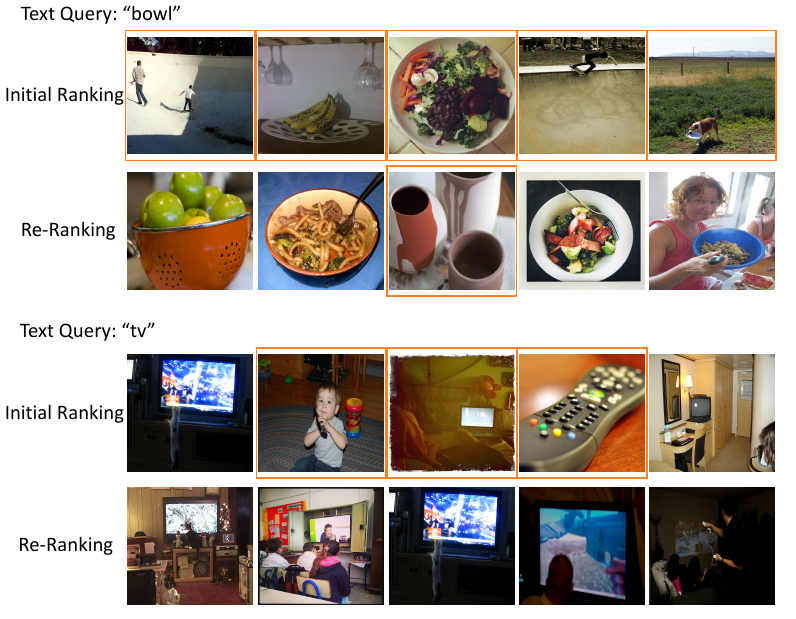}
	\caption{
\textbf{Qualitative Comparison on Occluded COCO for SigLIP and ELIP-S (top), SigLIP-2 and ELIP-S-2 (bottom).} 
Similar to Figure~\ref{fig:supple_qualitative_clip2-2}, we display examples at the top-100 rankings where there is a difference between the two models, \emph{i.e.,} one model retrieves a positive sample while the other model retrieves a negative sample. \emph{Generally, ELIP-S/ELIP-S-2 retrieves more positive samples in top-100 images than SigLIP/SigLIP-2}. Negative samples (errors) are highlighted in an orange solid box. 
For the first example, the text query is `bowl' and SigLIP confuses it with `roller skating rink', `plate', and `frisbee' in the top-100 retrieved images, while ELIP-S confuses it with `cup'; for the second example, the text query is `TV' and SigLIP-2 confuses it with `cell phone', 'laptop' and 'remote control' in top-100 retrieved images.}

	\label{fig:supple_qualitative_siglip2}
	\end{figure*}

\clearpage
Figure~\ref{fig:supple_qualitative_siglip2} displays qualitative comparison of SigLIP/SigLIP-2 initial ranking and our ELIP-S/ELIP-S-2 re-ranking on Occluded COCO.

\begin{figure*}[h]
	\centering
\includegraphics[height=0.75\linewidth]{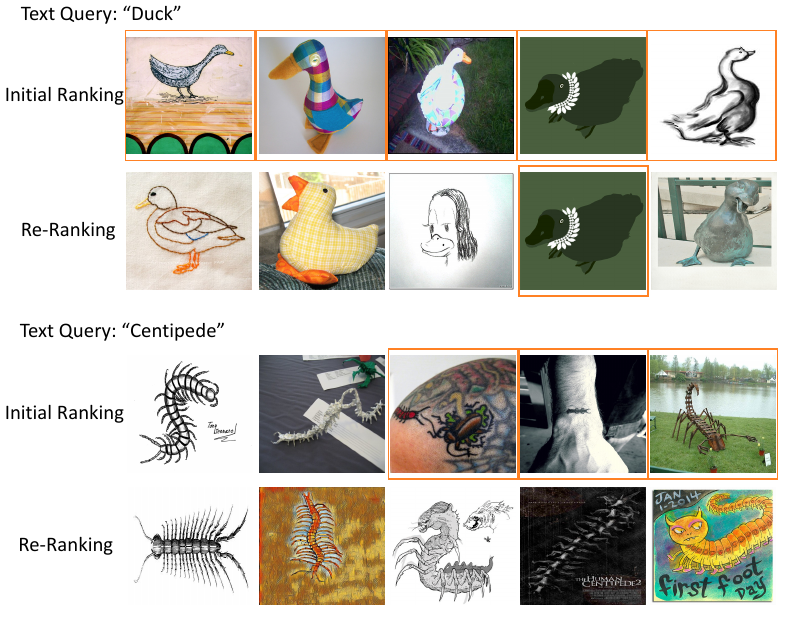}
	\caption{
\textbf{Qualitative Comparison on ImageNet-R for SigLIP and ELIP-S (top), SigLIP-2 and ELIP-S-2 (bottom).} We display examples at the top-100 rankings where there is a difference between the two models, \emph{i.e.,} one model retrieves a positive sample while the other model retrieves a negative sample, as the majority of the top-100 retrieves samples are positive. \emph{Generally, ELIP-S/ELIP-S-2 retrieves more positive samples in top-100 images than SigLIP/SigLIP-2}. Negative samples (errors) are highlighted in an orange solid box. 
For the first example, the text query is `duck' and SigLIP retrieves some `goose's in top-100 retrieved images while our ELIP-S retrieves a `goose' in top-100 images; for the second example, the text query is `centipede' and SigLIP-2 retrieves some other insects in top-100 retrieved images.  
	} 
	\label{fig:supple_qualitative_siglip2-2}
	\end{figure*}

\clearpage
Figure~\ref{fig:supple_qualitative_siglip2-2} displays qualitative comparison of SigLIP/SigLIP-2 initial ranking and our ELIP-S/ELIP-S-2 re-ranking on ImageNet-R.

\clearpage
In conclusion, these results demonstrate that, though the initial SigLIP/SigLIP-2 model can roughly retrieve relevant images, it is less able to distinguish fine-grained differences between the top retrieved images, whereas our re-ranking model further improves the discrimination between the ground truth and hard negative images, achieving a better retrieval.

\clearpage
\subsection{Qualitative Comparison of BLIP and ELIP-B}

\begin{figure*}[h]
	\centering
\includegraphics[height=0.75\linewidth]{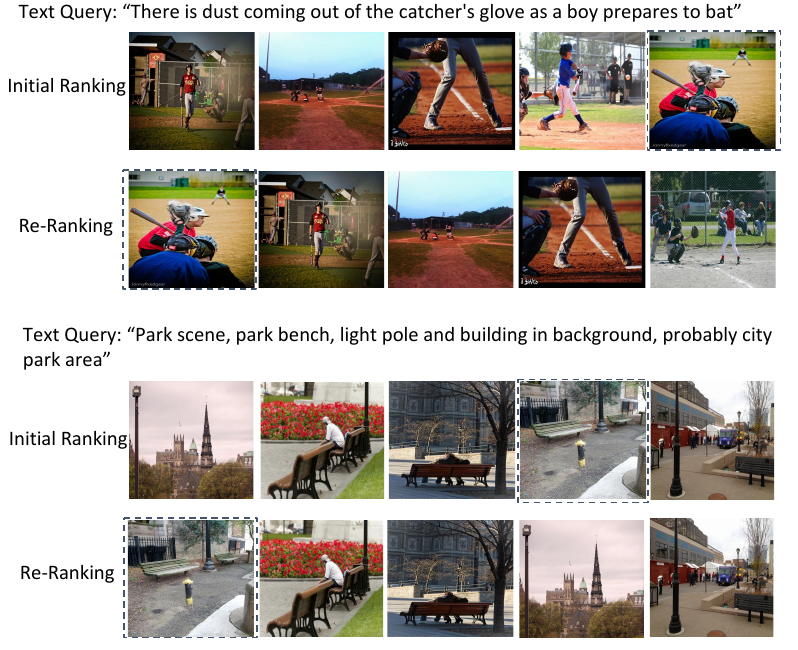}
	\caption{
\textbf{Qualitative Comparison on COCO for BLIP-2 and ELIP-B.} For each example, we show both top-5 rankings (from left to right) and highlight the ground truth image in a black dashed box.
In the first and second examples, the text queries are `There is dust coming out of the catcher's glove as a boy prepares to bat' and `Park scene, park bench, light pole and building in background, probably city park area', and the ground truth images are ranked top-5 / top-4 respectively by BLIP-2, but ranked top-1 by ELIP-B. 
	} 
	\label{fig:supple_qualitative_blip1}
	\end{figure*}

Figure~\ref{fig:supple_qualitative_blip1} shows more qualitative comparison between the BLIP-2 ranking and ELIP-B re-ranking on COCO. 

\begin{figure*}[h]
	\centering
\includegraphics[height=0.75\linewidth]{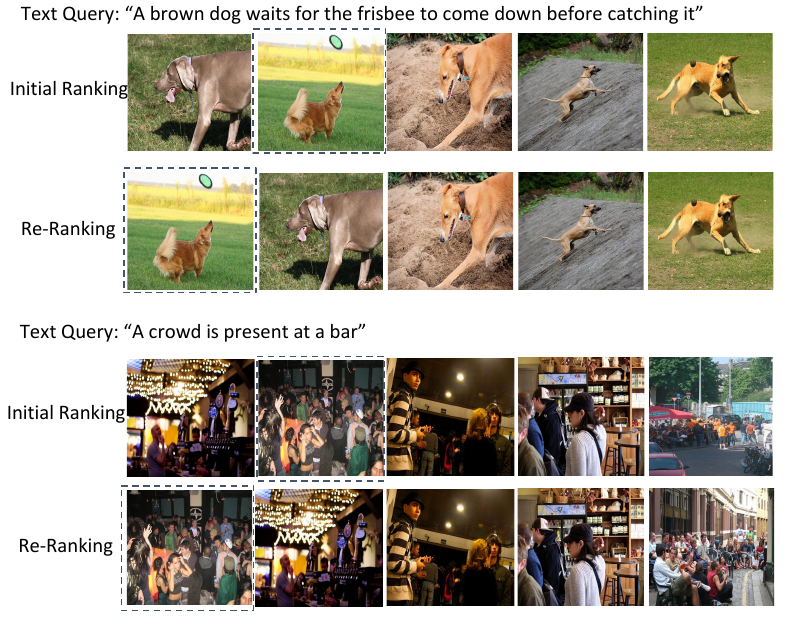}
	\caption{
\textbf{Qualitative Comparison on Flickr for BLIP-2 and ELIP-B.} For each example, we show both top-5 rankings (from left to right) and highlight the ground truth image in a black dashed box.
In the first and second examples, the text queries are `A brown dog waits for the frisbee to come down before catching it' and `A crowd is present at a bar', and the ground truth images rank top-2 in the BLIP-2 ranking but top-1 in our ELIP-B ranking.
	} 
	\label{fig:supple_qualitative_blip1-2}
	\end{figure*}

\clearpage
Figure~\ref{fig:supple_qualitative_blip1-2} shows more qualitative comparison between the BLIP-2 ranking and ELIP-B re-ranking on Flickr. 

\begin{figure*}[h]
	\centering
\includegraphics[height=0.75\linewidth]{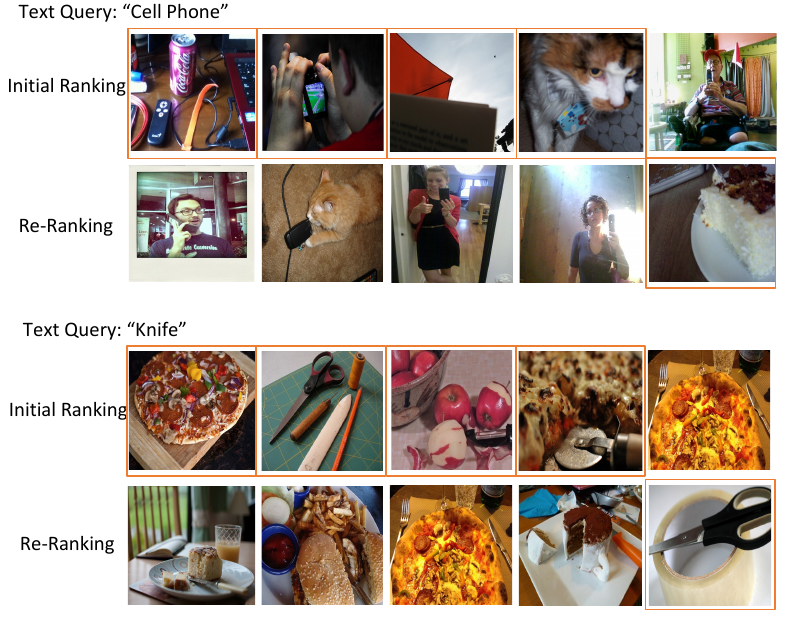}
	\caption{
\textbf{Qualitative Comparison on Occluded COCO for BLIP-2 and ELIP-B.} Similar to Figure~\ref{fig:supple_qualitative_clip2-2}, we display examples at the top-100 rankings where there is a difference between the two models, \emph{i.e.,} one model retrieves a positive sample while the other model retrieves a negative sample. \emph{Generally, ELIP-B retrieves more positive samples in top-100 images than BLIP-2}. Negative samples (errors) are highlighted in an orange solid box. 
For the first example, the text query is `cell phone' and BLIP-2 confuses it with `remote control' and `camera' in the top-100 retrieved images, while ELIP-B confuses it with `keyboard'; for the second example, the text query is `knife' and BLIP-2 confuses it with `scissors', `fruit peeler' and `pizza cutter' in top-100 retrieved images, while ELIP-B confuses it with `scissors'.  
	} 
	\label{fig:supple_qualitative_blip2}
	\end{figure*}

\clearpage
Figure~\ref{fig:supple_qualitative_blip2} displays qualitative comparison of BLIP-2 initial ranking and our ELIP-B re-ranking on Occluded COCO.

\begin{figure*}[h]
	\centering
\includegraphics[height=0.75\linewidth]{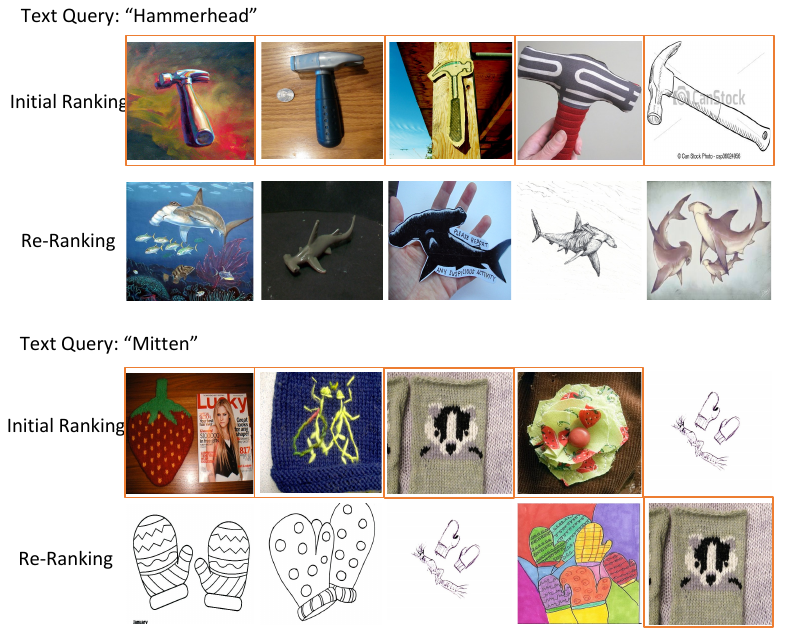}
	\caption{
\textbf{Qualitative Comparison on ImageNet-R for BLIP-2 and ELIP-B.} Similarly, we display examples at the top-100 rankings where there is a difference between the two models, \emph{i.e.,} one model retrieves a positive sample while the other model retrieves a negative sample. \emph{Generally, ELIP-B retrieves more positive samples in top-100 images than BLIP-2}. Negative samples (errors) are highlighted in an orange solid box. 
For the third example, the text query is `hammerhead' and BLIP-2 retrieves some `hammer's in the top-100 retrieved images; for the fourth example, the text query is `mitten' and BLIP-2 retrieves some other objects made of similar material of mitten in top-100 retrieved images while our ELIP-B makes a similar mistake but the total number of errors are lower. 
	} 
	\label{fig:supple_qualitative_blip2-2}
	\end{figure*}

\clearpage
Figure~\ref{fig:supple_qualitative_blip2-2} displays qualitative comparison of BLIP-2 initial ranking and our ELIP-B re-ranking on ImageNet-R.

\clearpage
In conclusion, these results demonstrate that the ELIP-B re-ranking model improves the discrimination capability between the positive and hard negative images, achieving a better retrieval performance than the original BLIP-2 model.

\clearpage
\subsection{Attention Map of CLIP and ELIP-C}

\begin{figure*}[h]
	\centering
\includegraphics[height=0.55\linewidth]{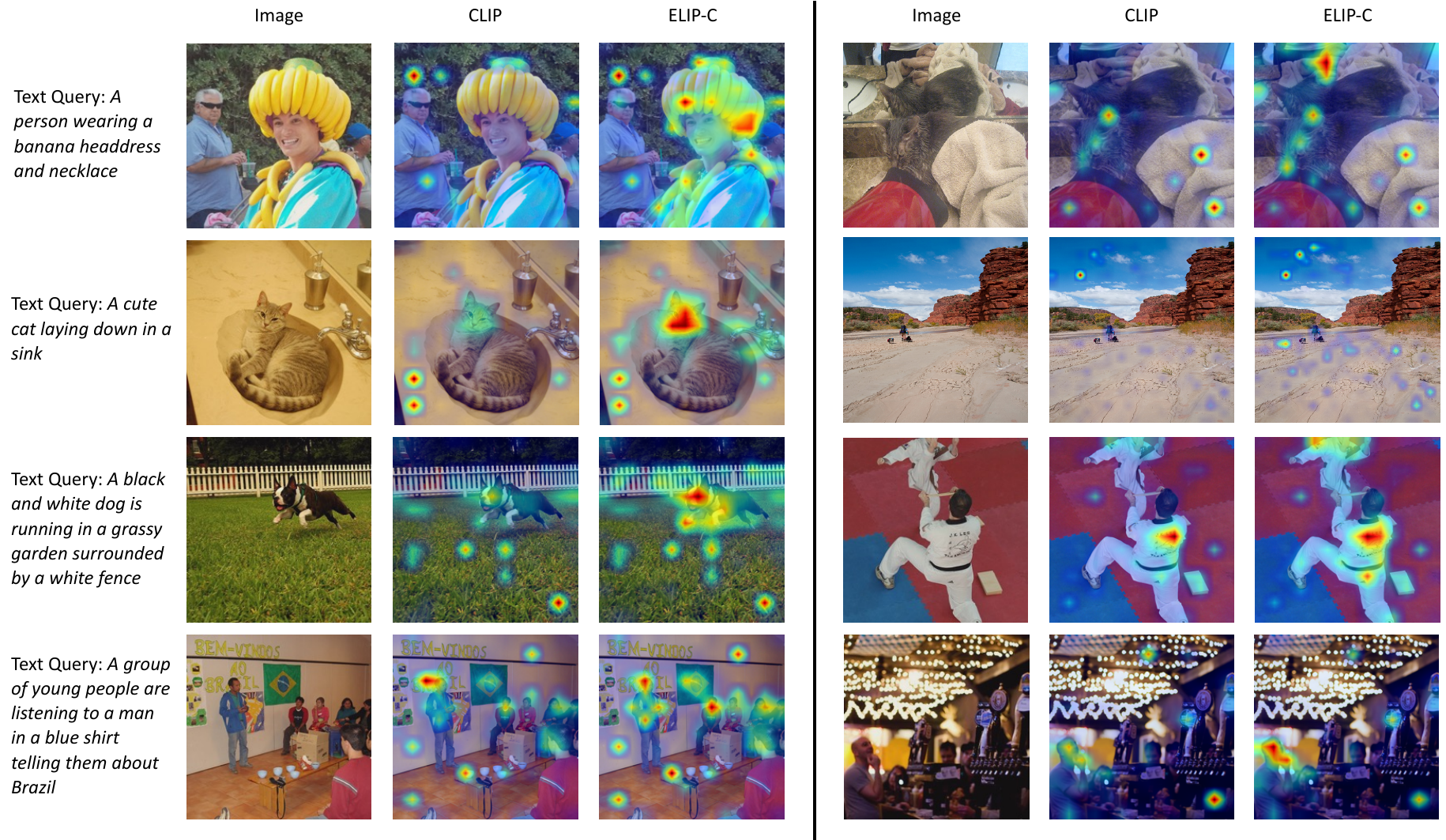}
	\caption{
\textbf{Visualisation of attention map of CLIP and ELIP-C} shows the cross-attention map of \texttt{[CLS]} token on patch tokens for both CLIP and our ELIP-C on COCO and Flickr. Left: image matches the text query; Right: image does not match the text query. COCO: Rows 1-2; Flickr: Rows 3-4.
It can be observed that when the image matches the text query, our generated visual prompt vectors can effectively boost the selection of image features relevant to the text query: 
For example, when the text query is `A person wearing a banana headdress and necklace'~(Row 1), comparing the attention map of CLIP and ELIP-C, we can observe ELIP-C enables more attention on the banana headdress and necklace; when the text query is `A cute cat laying down in a sink'~(Row 2), we note that ELIP-C gives significantly more attention on the cat; when the text query is `A black and white dog is running in a grassy garden surrounded by a white fence'~(Row 3), it can be noted that ELIP-C largely increases the attention on the black and white dog and the white fence; when the text query is `A group of young people are listening to a man in a blue shirt telling them about Brazil'~(Row 4), it can be noted that ELIP-C makes the attention to focus more on the man in a blue shirt who is talking, as well as the Brazil flag and people listening.
The difference is not significant if the image does not match the query (Columns 4-6). 
	} 
	\label{fig:supple_attention_clip}
	\end{figure*}

Figure~\ref{fig:supple_attention_clip} shows the cross-attention map of \texttt{[CLS]} token on patch tokens for both CLIP and our ELIP-C on COCO (Rows 1-2) and Flickr (Rows 3-4). We display the cases where the image matches the text query on the left and the images that do not match the text query on the right. The visualisation is based on the codebase of~\cite{selvaraju2017grad}, where a warmer color represents a higher activation.

\begin{figure*}[h]
	\centering
\includegraphics[height=0.55\linewidth]{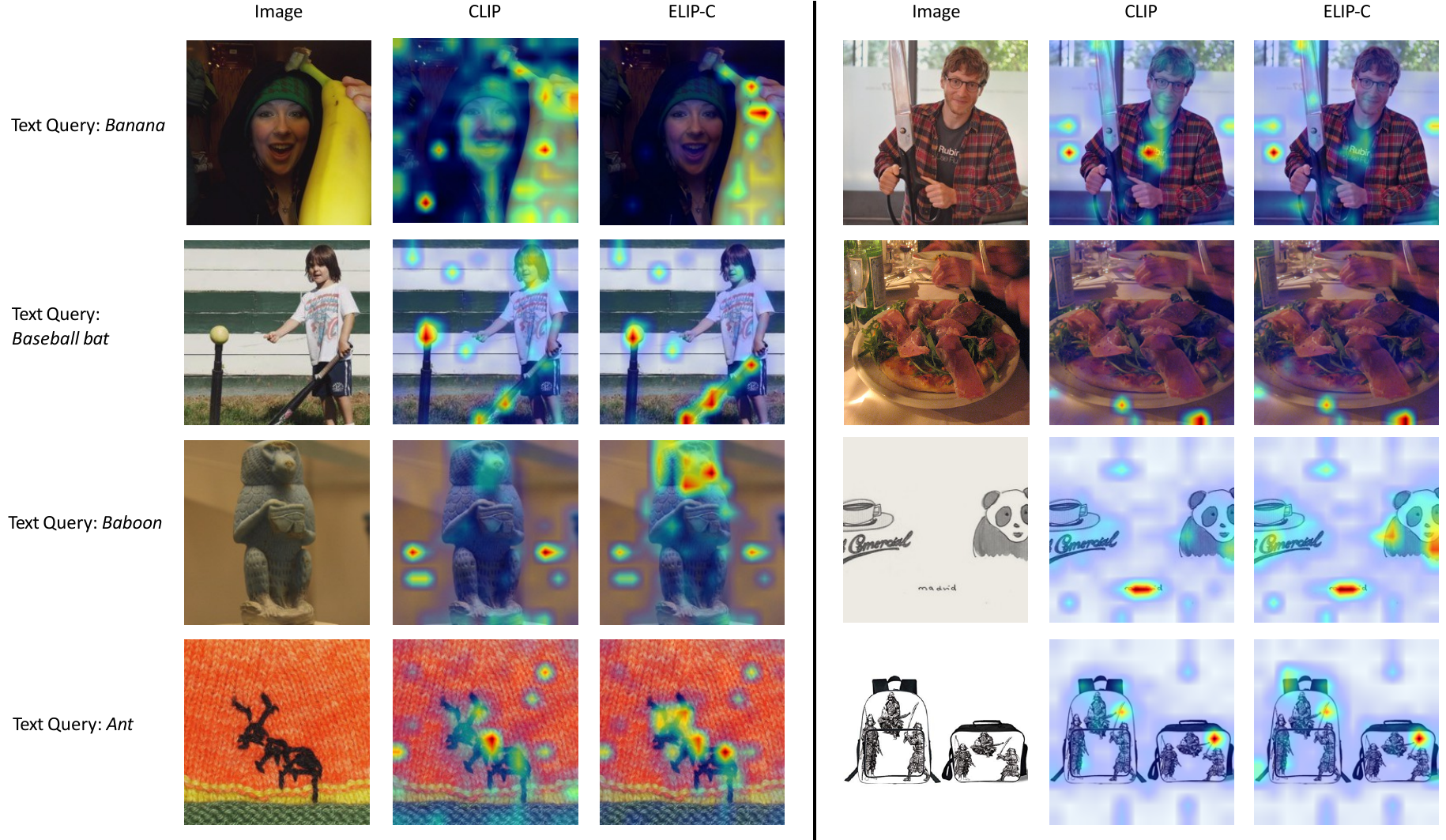}
	\caption{
\textbf{Visualisation of attention map of CLIP and ELIP-C} shows the cross-attention map of \texttt{[CLS]} token on patch tokens for both CLIP and our ELIP-C on Occluded COCO and ImageNet-R. Left: image matches the text query; Right: image does not match the text query. Occluded COCO: Rows 1-2; ImageNet-R: Rows 3-4.
It can be observed that when the image matches the text query, our generated visual prompt vectors can effectively boost the selection of image features relevant to the text query: 
For example, when the text query is `banana'~(Row 1), it can be observed that ELIP-C enables more concentration of attention on the banana rather than other areas; when the text query is `baseball bat'~(Row 2), we observe that ELIP-C brings more attention on the baseball bat; when the text query is `baboon'~(Row 3) and `ant'~(Row 4), ELIP-C makes the attention to focus more on the baboon and ant respectively.
The difference is not significant if the image does not match the query (Columns 4-6). 	} 
	\label{fig:supple_attention_clip-2}
	\end{figure*}

\clearpage
Figure~\ref{fig:supple_attention_clip-2} shows the cross-attention map of \texttt{[CLS]} token on patch tokens for both CLIP and our ELIP-C on Occluded COCO (Rows 1-2) and ImageNet-R (Rows 3-4). We display the cases where the image matches the text query on the left and the images that do not match the text query on the right. The visualisation is based on the codebase of~\cite{selvaraju2017grad}, where a warmer color represents a higher activation.

\clearpage
\subsection{Attention Map of SigLIP/SigLIP-2 and ELIP-S/ELIP-S-2}

\begin{figure*}[h]
	\centering
\includegraphics[height=0.55\linewidth]{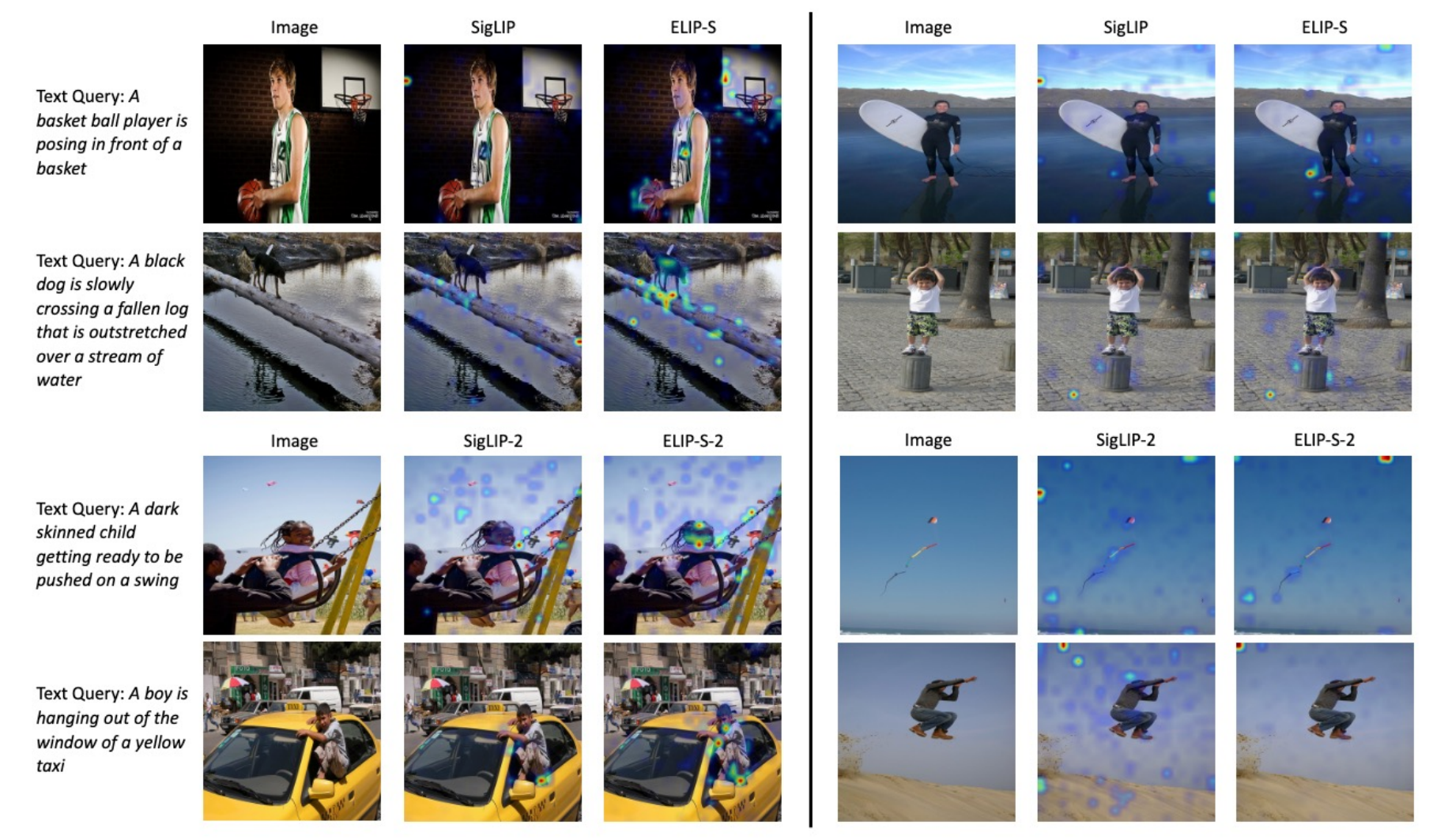}
	\caption{
\textbf{Visualisation of attention map of SigLIP/SigLIP-2 and ELIP-S/ELIP-S-2} shows the cross-attention map of \texttt{[CLS]} token on patch tokens for both SigLIP/SigLIP-2 and our ELIP-S/ELIP-S-2 on COCO and Flickr. Top: SigLIP v.s. ELIP-S; Bottom: SigLIP-2 v.s. ELIP-S-2. Left: image matches the text query; Right: image does not match the text query. COCO: Rows 1 and 3; Flickr: Rows 2 and 4. 
It can be observed that when the image matches the text query, our generated visual prompt vectors can effectively boost the selection of image features relevant to the text query: 
For example, when the text query is `A basket ball player is posing in front of a basket'~(Row 1), comparing the attention map of SigLIP and ELIP-S, we can observe ELIP-S enables more attention on the basketball, the basket, and the player; when the text query is `A black dog is slowly crossing a fallen log that is outstretched over a stream of water'~(Row 2), we note that ELIP-S gives significantly more attention on the black dog and the log; when the text query is `A dark skinned child getting ready to be pushed on a swing'~(Row 3), it can be noted that ELIP-S-2 largely increases the attention on the child and the swing; when the text query is `A boy is hanging out of the window of a yellow taxi'~(Row 4), it can be noted that ELIP-S-2 makes the attention to focus more on the boy out of the window.
The difference is not significant if the image does not match the query (Columns 4-6). 
	} 
	\label{fig:supple_attention_siglip}
	\end{figure*}

Figure~\ref{fig:supple_attention_siglip} shows the cross-attention map of \texttt{[CLS]} token on patch tokens for both SigLIP/SigLIP-2 and our ELIP-S/ELIP-S-2 on COCO (Rows 1 and 3) and Flickr (Rows 2 and 4). We display the cases where the image matches the text query on the left and the images that do not match the text query on the right. 
The visualisation is based on the codebase of~\cite{selvaraju2017grad}, where a warmer color represents a higher activation.

\begin{figure*}[h]
	\centering
\includegraphics[height=0.55\linewidth]{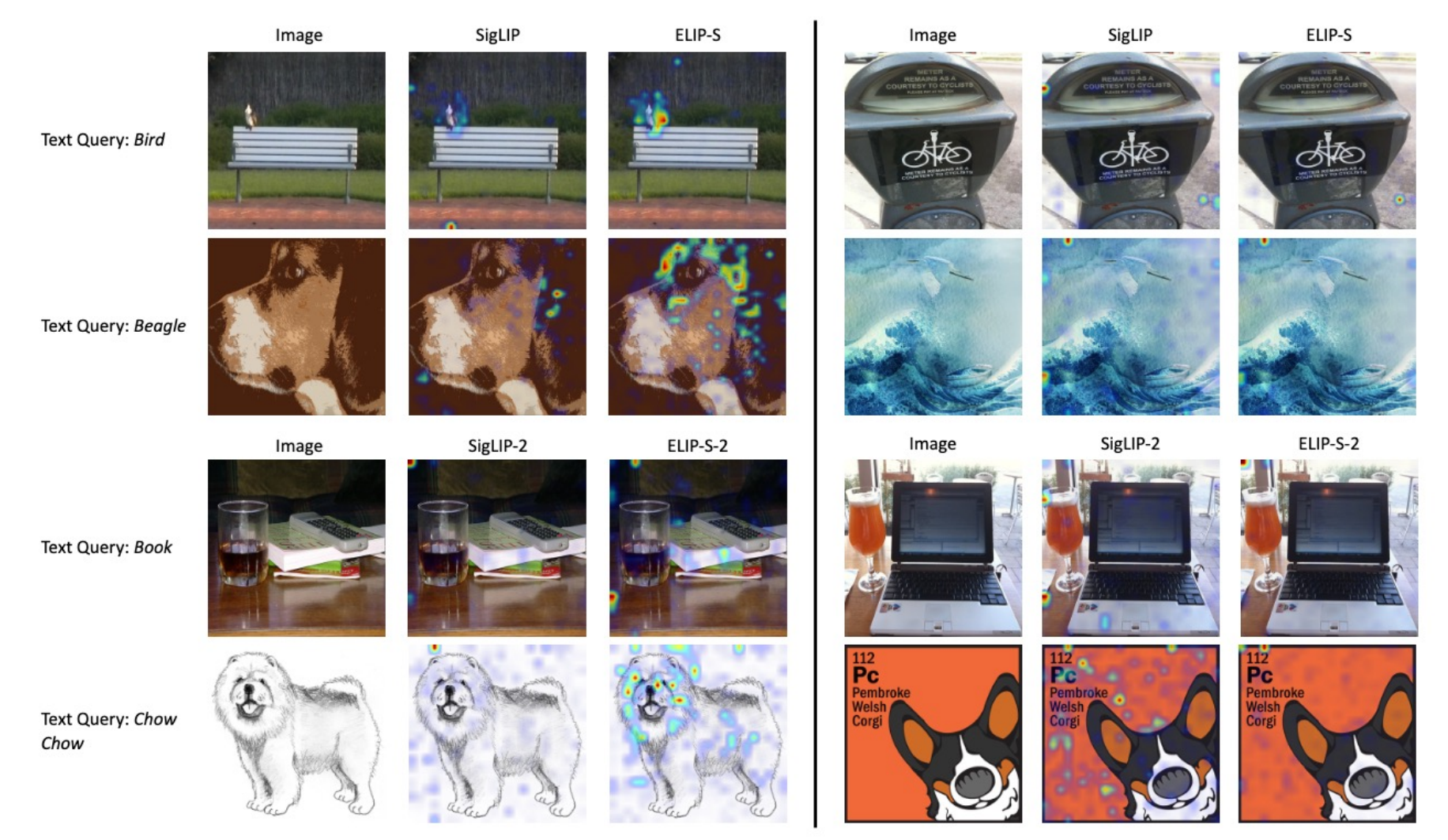}
	\caption{
\textbf{Visualisation of attention map of SigLIP/SigLIP-2 and ELIP-S/ELIP-S-2} shows the cross-attention map of \texttt{[CLS]} token on patch tokens for both SigLIP/SigLIP-2 and our ELIP-S/ELIP-S-2 on Occluded COCO and ImageNet-R. Top: SigLIP v.s. ELIP-S; Bottom: SigLIP-2 v.s. ELIP-S-2. Left: image matches the text query; Right: image does not match the text query. Occluded COCO: Rows 1 and 3; ImageNet-R: Rows 2 and 4.
It can be observed that when the image matches the text query, our generated visual prompt vectors can effectively boost the selection of image features relevant to the text query: 
For example, when the text query is `bird'~(Row 1), it can be observed that ELIP-S enables more concentration of attention on the bird; when the text query is `beagle'~(Row 2), we observe that ELIP-S brings more attention on the beagle; when the text query is `book'~(Row 3) and `chow chow'~(Row 4), ELIP-S-2 makes the attention to focus more on the book and chow chow respectively.
The difference is not significant if the image does not match the query (Columns 4-6). 
	} 
	\label{fig:supple_attention_siglip-2}
	\end{figure*}

\clearpage
Figure~\ref{fig:supple_attention_siglip-2} shows the cross-attention map of \texttt{[CLS]} token on patch tokens for both SigLIP/SigLIP-2 and our ELIP-S/ELIP-S-2 on Occluded COCO (Rows 1 and 3) and ImageNet-R (Rows 2 and 4). We display the cases where the image matches the text query on the left and the images that do not match the text query on the right. 
The visualisation is based on the codebase of~\cite{selvaraju2017grad}, where a warmer color represents a higher activation.

\clearpage
\subsection{Attention Map of BLIP-2 and ELIP-B}

\begin{figure*}[h]
	\centering
\includegraphics[height=0.55\linewidth]{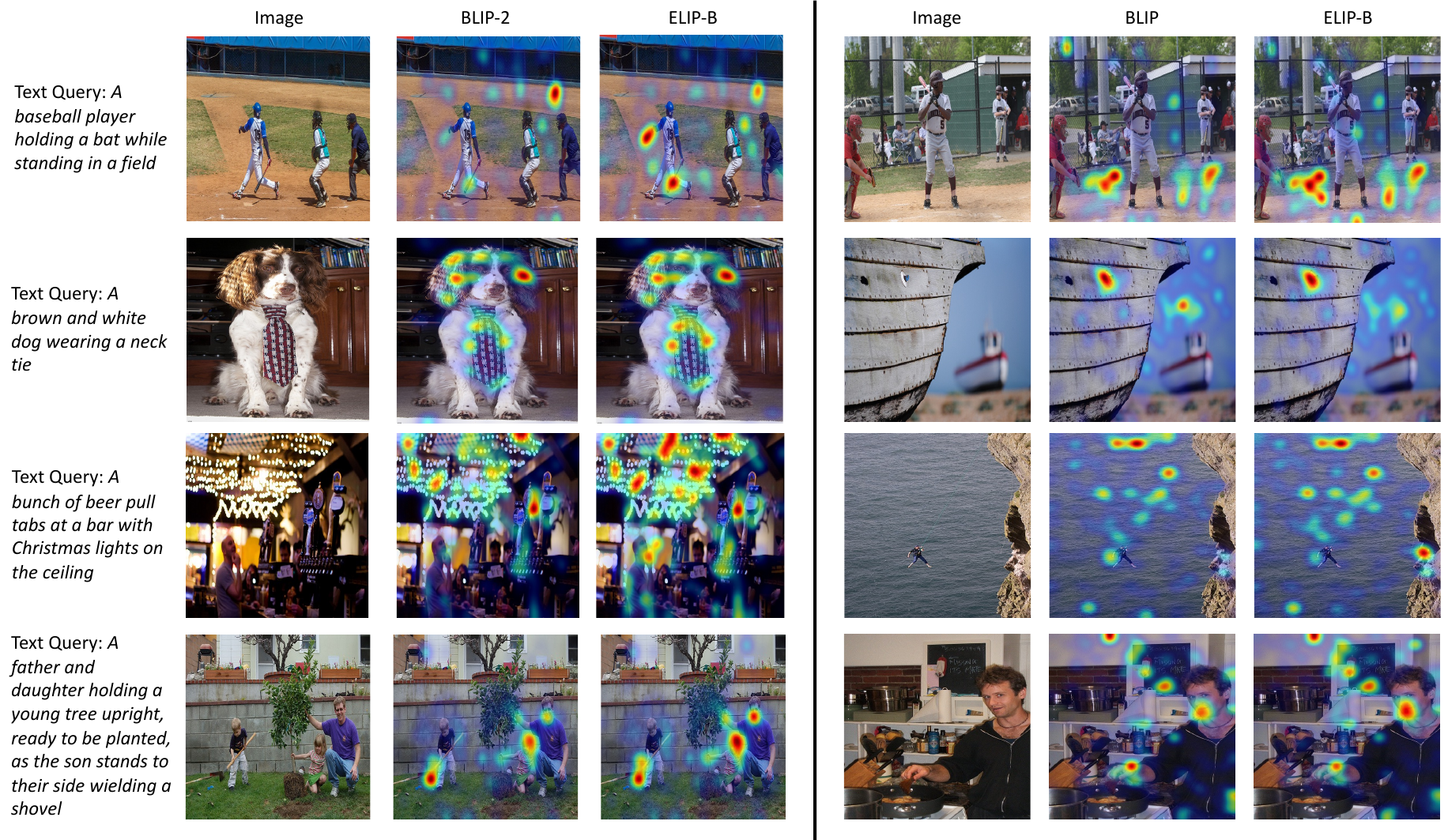}
	\caption{
\textbf{Visualisation of attention map of BLIP-2 and ELIP-B} shows the cross-attention map of the query tokens at the ITM head on patch tokens for both BLIP-2 and our ELIP-B on COCO and Flickr. There are 32 query tokens in total, and we take the average value of the 32 query tokens to visualise the attention map. The attention map represents image features at which locations are selected and concentrated.
Left: image matches the text query; Right: image does not match the text query. COCO: Rows
1-2; Flickr: Rows 3-4. 
It can be observed that when the image matches the text query, our generated visual prompt vectors can effectively boost the selection of image features relevant to the text query: 
For example, when the text query is `A baseball player holding a bat while standing in a field'~(Row 1), comparing the attention map of BLIP-2 and ELIP-B we can observe ELIP-B enables more attention on the baseball bat; 
when the text query is `A brown and white dog wearing a neck tie'~(Row 2), we note that ELIP-B gives more attention on the tie and the dog; 
when the text query is `A bunch of beer pull tabs at a bar with Christmas lights on the ceiling'~(Row 3), it can be noted that ELIP-B largely increases the attention on the Christmas lights on the ceiling and the bunch of beer pull tabs; 
when the text query is `A father and daughter holding a young tree upright, ready to be planted, as the son stands to their side wielding a shovel'~(Row 4), it can be noted that ELIP-B makes the attention focus more on the father, daughter, son and tree.  
The difference is not significant if the image does not match the query (Columns 4-6). 
	} 
	\label{fig:supple_attention_blip}
	\end{figure*}

Figure~\ref{fig:supple_attention_blip} shows the cross-attention map of the query tokens at the ITM head on patch tokens for both BLIP-2 and our ELIP-B on COCO (Rows 1-2) and Flickr (Rows 3-4). There are 32 query tokens in total, and we take the average value of the 32 query tokens to visualise the attention map. The attention map represents image features at which locations are selected and concentrated. We display the situations where the image matches the text query on the left, and where the images do not match the text query on the right. The visualisation is based on the codebase of~\cite{selvaraju2017grad}, where a warmer color represents a higher activation.

\begin{figure*}[h]
	\centering
\includegraphics[height=0.55\linewidth]{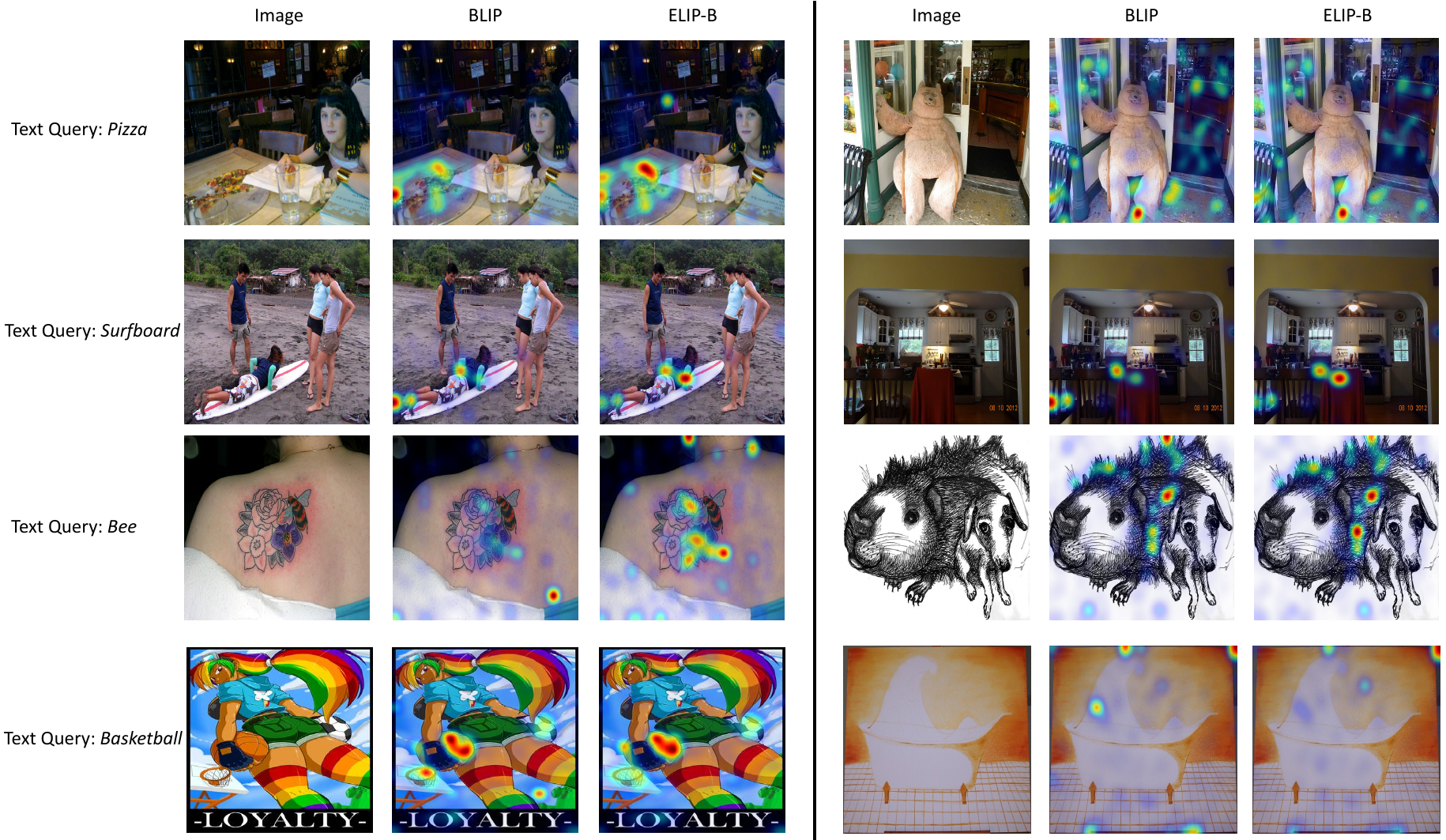}
	\caption{
\textbf{Visualisation of attention map of BLIP-2 and ELIP-B} shows the cross-attention map of the query tokens at the ITM head on patch tokens for both BLIP-2 and our ELIP-B on Occluded COCO and ImageNet-R. There are 32 query tokens in total, and we take the average value of the 32 query tokens to visualise the attention map. The attention map represents image features at which locations are selected and concentrated.
Left: image matches the text query; Right: image does not match the text query. Occluded COCO: Rows 1-2; ImageNet-R: Rows 3-4. It can be observed that when the image matches the text query, our generated visual prompt vectors can effectively boost the selection of image features relevant to the text query: 
For example, when the text query is `pizza'~(Row 1), it can be observed that ELIP-B enables more concentration of attention on the pizza; 
when the text query is `surfboard'~(Row 2), we observe that ELIP-B brings more attention on the surfboard; 
when the text query is `bee'~(Row 3) and `basketball'~(Row 4), ELIP-B makes the attention to focus more on the bee and basketball respectively. 
The difference is not significant if the image does not match the query (Columns 4-6). 
	} 
	\label{fig:supple_attention_blip-2}
	\end{figure*}

\clearpage
Figure~\ref{fig:supple_attention_blip-2} shows the cross-attention map of the query tokens at the ITM head on patch tokens for both BLIP-2 and our ELIP-B on Occluded COCO (Rows 1-2) and ImageNet-R (Rows 3-4). There are 32 query tokens in total, and we take the average value of the 32 query tokens to visualise the attention map. The attention map represents image features at which locations are selected and concentrated. We display the situations where the image matches the text query on the left and where the images do not match the text query on the right. The visualisation is based on the codebase of~\cite{selvaraju2017grad}, where a warmer color represents a higher activation.

\clearpage
\section{Ablation Study for Different Number of Generated Visual Prompt Tokens}
\label{sec:ablation_different_number_vpt_tokens}

\begin{table*}[h]
    \centering
    \tabcolsep=0.25cm
    \begin{tabular}{ccccccccc}
    \toprule
 \multirow{2}*{Settings} & \multirow{2}*{\#Generated Tokens} & \multicolumn{3}{c}{COCO} &  \multicolumn{3}{c}{Flickr}& \\
\cmidrule(lr){3-5} \cmidrule(lr){6-8}
& &  Recall@1 & Recall@5 & Recall@10  & Recall@1 & Recall@5 & Recall@10 \\
\midrule

$\mathbb{A}$ & 0 & 40.2 & 66.0 & 75.6 & 67.6 & 88.3 & 93.0 \\ 
$\mathbb{B}$ & 1 & 44.2 & 70.0 & 79.5 & 71.3 & 90.6 & 94.4  \\ 
$\mathbb{C}$ & 2 & 44.7 & 70.2 & 79.8 & 71.4 & 90.5 & 94.1  \\ 
$\mathbb{D}$ & 5 & 45.2 & 70.5 & 80.0 & 72.4 & 90.5 & 94.6  \\ 
$\mathbb{E}$ & 10 & 45.6 & 71.1 & 80.4 & 72.3 & 90.6 & 94.7  \\

    \bottomrule
    \end{tabular}
    \caption{\textbf{Ablation study on number of generated visual prompt tokens.} Following Table~1 of the main paper, we use CLIP as the baseline for the ablation study. 0 refers to the CLIP baseline and a non-zero number refers to our ELIP-C architecture with different numbers of generated visual prompt vectors.}
    \label{tab:ablation_num_token}
\end{table*}

As mentioned in Section~7.1, here we study the performance as the number of generated visual prompt tokens are varied. The results are shown in Table~\ref{tab:ablation_num_token}, and it can be observed that generating multiple tokens, \emph{e.g.,} 2 tokens (Setting $\mathbb{C}$), 5 tokens (Setting $\mathbb{D}$) or 10 tokens (Setting $\mathbb{E}$), brings an improvement compared with generating only 1 token (Setting $\mathbb{B}$), and Setting $\mathbb{E}$ performs best. Therefore, we generate 10 tokens in all experiments.

\clearpage
\section{Further Ablation Study for ELIP-B}
\label{sec:sup_ablation_elip_b}

\begin{table*}[h]
    \centering
    \tabcolsep=0.1cm
    \begin{tabular}{cccccccccc}
    \toprule
 \multirow{2}*{Settings} &\multirow{2}*{Fine-tune ITM Head} & \multirow{2}*{JEST Selection} & \multicolumn{3}{c}{COCO} &  \multicolumn{3}{c}{Flickr}& \\
\cmidrule(lr){4-6} \cmidrule(lr){7-9}
& & & Recall@1 & Recall@5 & Recall@10  & Recall@1 & Recall@5 & Recall@10 \\
\midrule

$\mathbb{A}$ & - & - & 68.25 & 87.72 & 92.63 & 89.74 & 98.18 & 98.94 \\ 
$\mathbb{B}$ & \checkmark &  & 68.26 & 87.86 & 92.78 & 89.80 & 98.26 & 99.14 \\ 
$\mathbb{C}$ &  & \checkmark & 67.35 & 87.38 & 92.53 & 89.08 & 98.06 & 99.08 \\ 
$\mathbb{D}$ & \checkmark & \checkmark & \textbf{68.41} & \textbf{87.88} & \textbf{92.78} & \textbf{90.08} & \textbf{98.34} & \textbf{99.22} \\

    \bottomrule
    \end{tabular}
    \caption{\textbf{Ablation study of ELIP-B} in terms of \emph{training data selection inspired by JEST~\cite{evans2024data}} and \emph{fine-tuning ITM head}. Setting $\mathbb{A}$ is the baseline BLIP-2. Setting $\mathbb{B}$ is trained on a random subset of the same size of Settings $\mathbb{C}$ and $\mathbb{D}$. It can be observed that both the strategies of fine-tuning the ITM head and training on selected subset by JEST contribute to the improvement of our ELIP-B.}
    \label{tab:supple_blip_ablation}
\end{table*}

In this section, we have further conducted an ablation study on ELIP-B to show the effectiveness of selecting the training data using the strategy inspired by JEST~\cite{evans2024data} as described in Section~5.2, as well as also fine-tuning the ITM head as in Figure~3 of the main paper, as these are two unique things we only applied to BLIP-2.

In Table~\ref{tab:supple_blip_ablation}, we can observe that: 1) Comparing Settings $\mathbb{B}$ and $\mathbb{D}$, we can note the effectiveness of training on JEST selected training batches compared with selecting batches randomly; 2) Comparing Settings $\mathbb{C}$ and $\mathbb{D}$, we can infer that it is important to also fine-tune the lightweight ITM head as the input features to the ITM head have been changed in ELIP-B.

\clearpage
\section{Efficiency of ELIP Pre-Training}
\label{sec:sup_efficiency}

\begin{table}[h]
    \centering
    \tabcolsep=0.7cm
    \begin{tabular}{cccccc}
    \toprule 
    Model & GPU Hours & \#GPU & BS & FLOPS & $\Delta$FLOPS \\
    \midrule

CLIP & 10736 A100 & 176 $\times$ A100 & 33792 &33.7G & -  \\
ELIP-C  & 144 A40 & 2 $\times$  A40 & 40 &35.8G & 2.1G \\
\midrule
SigLIP & 1152 TPUv4 & 16 $\times$ TPUv4 & 32768 & 604.5G & -    \\
ELIP-S & 144 A40 & 2 $\times$  A40 & 10 &613.3G & 8.8G   \\
\midrule
SigLIP-2 & * & 2048 $\times$ TPUv5e & 32768 &1.31T & -    \\
ELIP-S-2 & 144 A40 & 2 $\times$  A40 & 10 &1.34T &  0.03T   \\
\midrule
BLIP-2 & 2304 A100 & 16 $\times$  A100 & 1680 & 1.33T & -   \\
ELIP-B & 48 A40 & 2 $\times$ A40 & 12 &1.35T & 0.02T \\

    \bottomrule
    \end{tabular}
    \caption{\textbf{Efficiency of ELIP pre-training} compared with original pre-training. Our method significantly improves the efficiency on training time, GPU requirement, and batch size, introducing only a marginal increase on FLOPS for our trainable MLP mapping network. *SigLIP-2 original training time is not provided in their paper.}
    \label{tab:supple_efficiency}
\end{table}

As mentioned in Section~7, in the table below we show a comparison of our method to the original pre-training of CLIP, SigLIP, SigLIP-2, and BLIP-2 in terms of training time, GPU requirement, batch size, and FLOPS. It can be observed that our method significantly improves the efficiency on training time, GPU requirement, and batch size, while only introducing a marginal increase on FLOPS which results from our trainable MLP mapping network.

\end{document}